\documentclass{article}

\PassOptionsToPackage{dvipsnames}{xcolor}
\usepackage[normalem]{ulem}
\usepackage{multirow}
\usepackage{bbm}
\usepackage{anyfontsize}
\usepackage{xspace}
\usepackage{changepage}
\usepackage[export]{adjustbox}
\usepackage{etoolbox}
\usepackage[table]{xcolor}
\usepackage{graphicx}
\usepackage{makecell} 
\usepackage{array} 
\usepackage{pifont}
\usepackage{enumitem}
\usepackage{svg}
\usepackage{tabularx}
\usepackage{subcaption} 

\newtoggle{citemain}
\iftoggle{citemain}{
\newcommand{\citeMain}[1]{\cite{#1}}
}{
\newcommand{\citeMain}[1]{{\color{red}\bf X}}
}

\definecolor{lightblue}{RGB}{219,234,254}
\definecolor{tealinc}{RGB}{13,148,136} 
\newcommand{\inc}[1]{\textcolor{tealinc}{#1}}
\definecolor{davemauve}{RGB}{180,130,200} 

\newcommand{\emoji}[2][3.0ex]{%
  \raisebox{-0.6ex}{\includegraphics[height=#1]{#2}}%
}

\definecolor{forestgreen}{rgb}{0.0, 0.50, 0.13}

\newcommand{\del}[1]{\iffalse #1 \fi} 

\newtoggle{comments}
\newtoggle{BSetal}

\toggletrue{comments}
\toggletrue{BSetal}

\newcommand{\myparagraph}[2][0]{\vspace{#1em}\noindent{\bf #2}}

\newcommand\bcos{B-cos\xspace}

\definecolor{hirow}{HTML}{F3E6D9}
\definecolor{hiblock}{HTML}{EFEFEF} 

\definecolor{mygray}{gray}{0.75} 

\newcommand{\RowStyle}{}
\newcolumntype{L}{>{\RowStyle}l}
\newcolumntype{R}{>{\RowStyle}r}

\usepackage[preprint]{icml2026}

\usepackage{microtype}
\usepackage{graphicx}
\usepackage{subcaption}
\usepackage{booktabs}

\usepackage{amsmath}
\usepackage{amssymb}
\usepackage{mathtools}
\usepackage{amsthm}

\usepackage[table]{xcolor}
\usepackage{colortbl}

\usepackage{hyperref}
\usepackage[capitalize,noabbrev]{cleveref}

\theoremstyle{plain}

\theoremstyle{definition}

\theoremstyle{remark}

\icmltitlerunning{DAVE: Distribution-aware Attribution via ViT Gradient Decomposition}

\begin{document}

\twocolumn[
  \icmltitle{DAVE~\texorpdfstring{\emoji{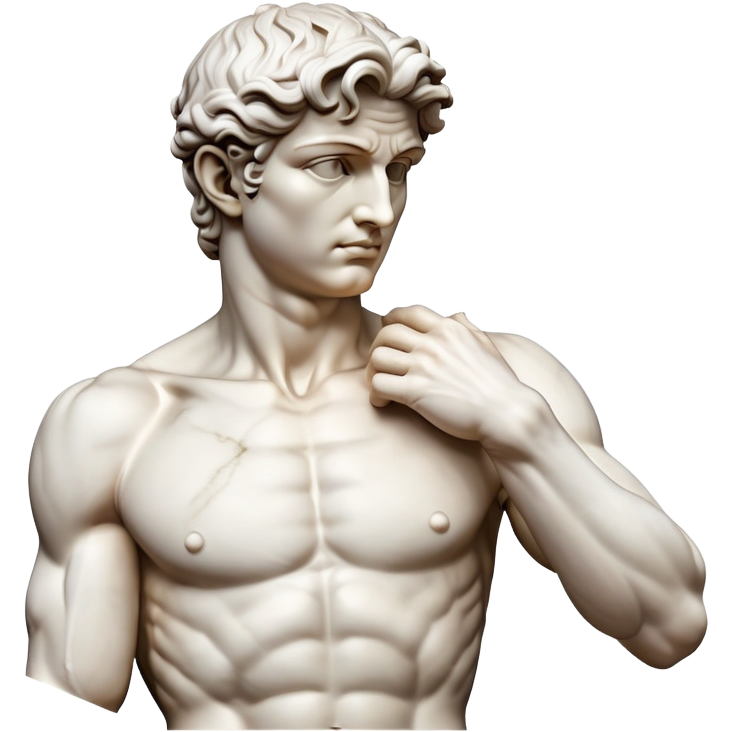}}{[dave]}: Distribution-aware Attribution via ViT Gradient Decomposition}
  
  \begin{icmlauthorlist}
    \icmlauthor{Adam Wróbel}{1,2}
    \icmlauthor{Siddhartha Gairola}{3}
    \icmlauthor{Jacek Tabor}{1}
    \icmlauthor{Bernt Schiele}{3}
    \icmlauthor{Bartosz Zieliński}{1}
    \icmlauthor{Dawid Rymarczyk}{1,4}
  \end{icmlauthorlist}

  \icmlaffiliation{1}{Jagiellonian University, Faculty of Mathematics and Computer Science}
  \icmlaffiliation{2}{Jagiellonian University, Doctoral School of Exact and Natural Sciences}
  \icmlaffiliation{3}{Max Planck Institute for Informatics, Saarland Informatics Campus, Saarbrucken, Germany}
  \icmlaffiliation{4}{Ardigen SA}

  \icmlcorrespondingauthor{Dawid Rymarczyk}{dawid.rymarczyk@uj.edu.pl}

  \icmlkeywords{Machine Learning, ICML}


  {%
\renewcommand\twocolumn[1][]{#1}%
\includegraphics[width=1.0\textwidth]{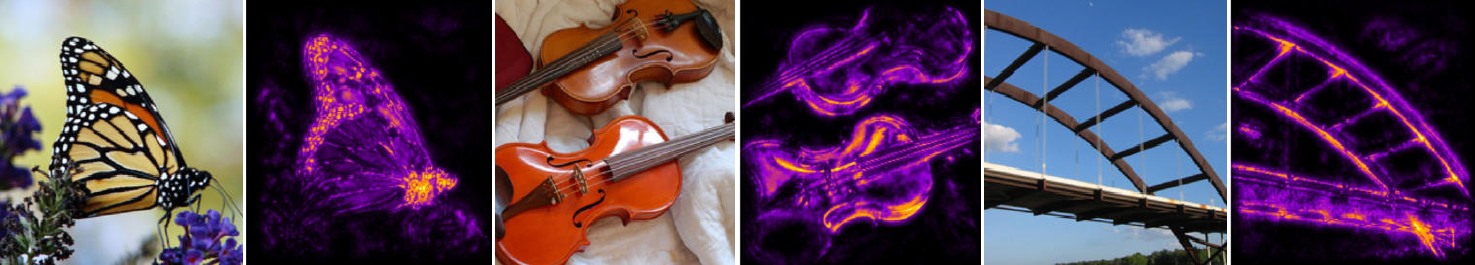}

\captionof{figure}{\textbf{DAVE provides fine-grained, pixel-level attributions that capture class-specific structural patterns present in each object. } \\ 
ImageNet-1k~\cite{imagenet} samples and corresponding DAVE attributions for DeiT III-B-16/224~\cite{touvron2022deit}.}

\vspace{2em}
\label{fig:teaser}
}

]



\printAffiliationsAndNotice{}  

\noindent
\begin{abstract}
Vision Transformers (ViTs) have become a dominant architecture in computer vision, yet producing stable and high-resolution attribution maps for these models remains challenging. Architectural components such as patch embeddings and attention routing often introduce structured artifacts in pixel-level explanations, causing many existing methods to rely on coarse patch-level attributions. We introduce DAVE \textit{(\underline{D}istribution-aware \underline{A}ttribution via \underline{V}iT Gradient D\underline{E}composition)}, a mathematically grounded attribution method for ViTs based on a structured decomposition of the input gradient. By exploiting architectural properties of ViTs, DAVE isolates locally equivariant and stable components of the effective input--output mapping. It separates these from architecture-induced artifacts and other sources of instability. 
Consequently, DAVE produces robust, precise and class-consistent attribution maps that reliably highlight visual features used by the model across inputs. Experimental results demonstrate that DAVE attributions are more stable and spatially precise than existing approaches.
\end{abstract}

\section{Introduction}

~\looseness=-1Deep neural networks have redefined the state of the art in computer vision~\cite{khan2022transformers}, yet their deployment in high-stakes applications such as medical diagnostics and autonomous driving remains limited by challenges in providing reliable and actionable decision explanations~\cite{rudin2019stop}. To address this, attribution methods in Explainable AI (XAI) have emerged~\cite{bach2015pixel,selvaraju2017grad} which aim to identify input features that most influence model predictions. While they are widely adopted for their model-agnostic nature, yet producing \textit{stable} and \textit{high-resolution} attributions has become increasingly challenging as modern architectures grow more complex~\cite{gairola2025probe}.

\begin{figure}
    \centering
    \includegraphics[width=1.0\linewidth]{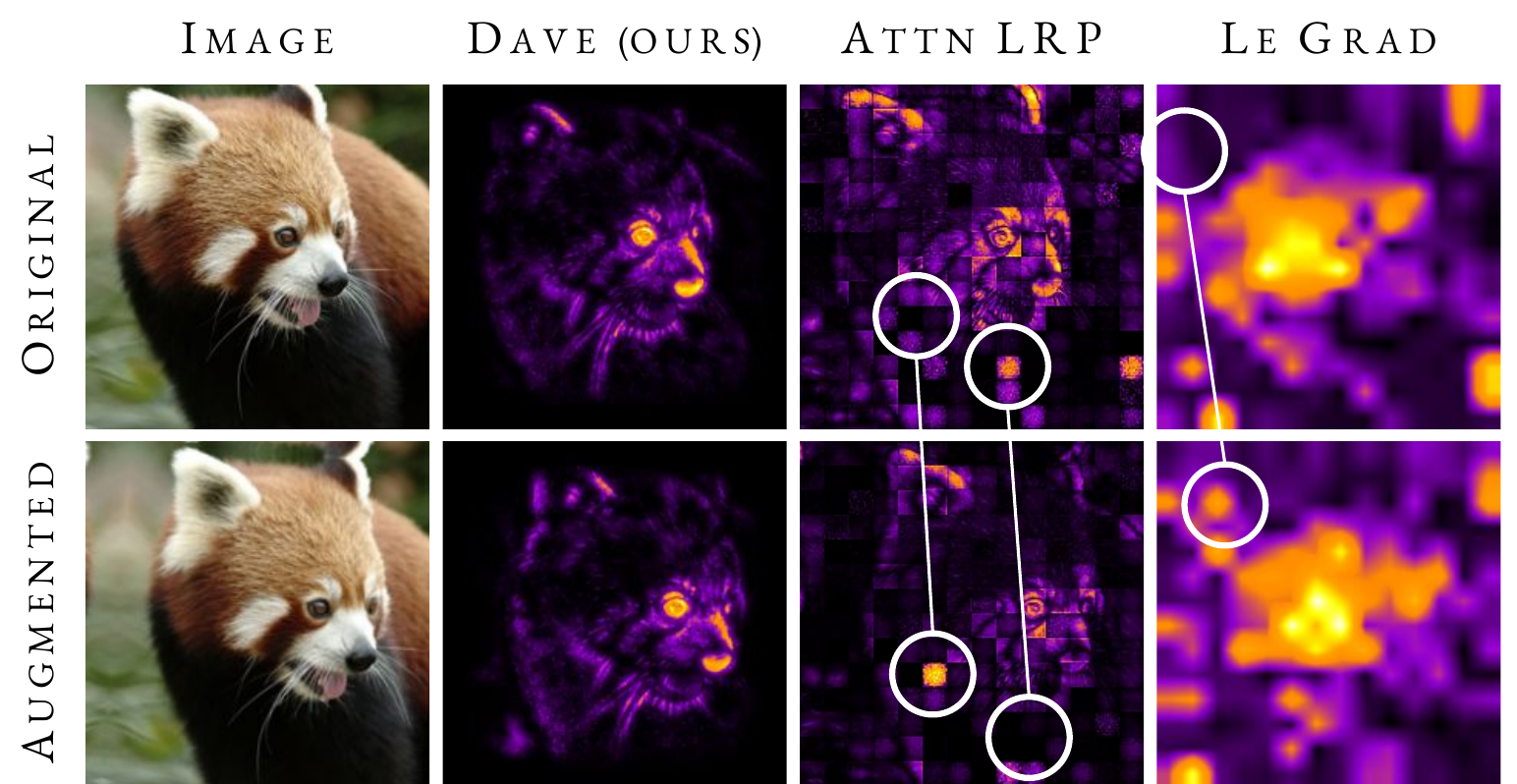}
    \caption{\textbf{Attribution consistency.} Under small augmentations (\textcolor{red!75}{5° rotation, 20px horizontal and 8px vertical shift}), DAVE highlights consistent features across the original and augmented images, while AttnLRP and LeGrad show inconsistent attributions (white markers), on a DeiT-III-B-16/224 model.}
    \label{fig:aug_example}
    \vspace{-2mm}
\end{figure}

~\looseness=-1Vision Transformers (ViTs)~\cite{dosovitskiy2020image} have become a dominant architecture in computer vision, achieving state-of-the-art performance across a wide range of tasks. Unlike CNNs, ViTs process images as sequences of patch tokens and rely on attention-based token mixing and learned projections. While effective for prediction, these architectural characteristics pose significant challenges for attribution methods~\cite{komorowski2023towards}. In particular, gradient- and attention-based explanations often exhibit structured, architecture-induced artifacts, leading either to unstable pixel-level attributions or to coarse patch-level explanations that lack fine-grained visual evidence (see Figure~\ref{fig:aug_example}). As a result, obtaining stable and high-resolution attributions for Vision Transformers remains an open challenge.

~\looseness-1To address these challenges, we introduce \textbf{DAVE}---\textbf{(\underline{D}istribution-aware \underline{A}ttribution via \underline{V}iT Gradient D\underline{E}composition)}, an attribution method suited for Vision Transformers that produces stable, high-resolution explanations. DAVE is based on a structured decomposition of the input gradient that leverages architectural properties of ViTs to identify locally equivariant and stable components of the model’s effective input–output transformation. By suppressing architecture-induced artifacts and separating transformation effects from unstable  input-dependent variations, DAVE enables robust pixel-level attribution without falling back to coarse patch-based explanations (see~\cref{fig:teaser}). As a result, the method highlights visual features that are consistently associated with model predictions across inputs and classes.


Our \textbf{contributions} can be summarized as follows:
\vspace{-2mm}
\begin{itemize}
    \item We propose a mathematically grounded attribution framework based on a structured decomposition of the input gradient that identifies a stable and locally equivariant component of the model’s effective input--output transformation.
    \item We apply this framework for Vision Transformers as \textbf{DAVE} (Distribution-aware Attribution via ViT Gradient Decomposition), which leverages ViT architectural properties to suppress patch- and attention-induced artifacts and produce stable, high-resolution attributions.
    \item We demonstrate DAVE’s effectiveness on multiple XAI benchmarks across both fully supervised and self-supervised foundation models (e.g., DeiT and DINO ViTs~\cite{touvron2021training,caron2021emerging}), and show its versatility by applying it to other architectures (e.g., inherently interpretable B-cos networks~\cite{bohle2022b}).
\end{itemize}

\begin{figure*}[!ht]
    \centering
    \includegraphics[width=1.0\textwidth]{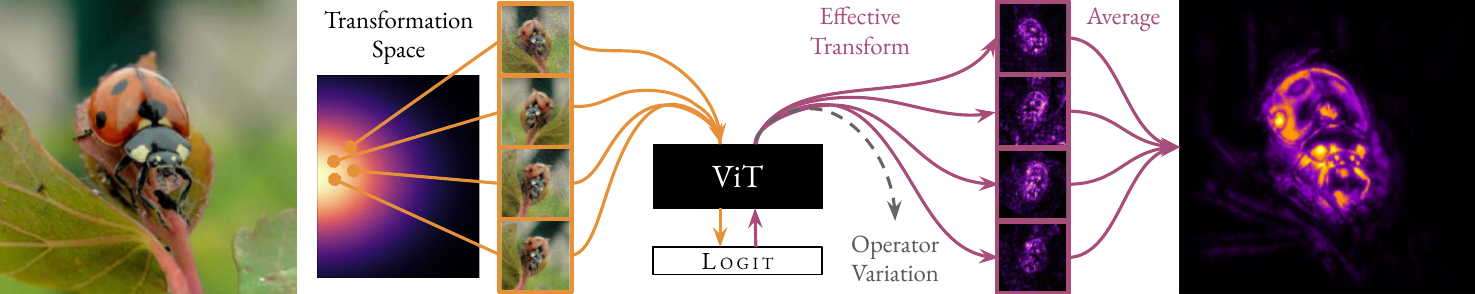}
   \caption{\textbf{Overview of the DAVE attribution pipeline for Vision Transformers.} 
Given an input image, DAVE samples small spatial transformations and Gaussian perturbations, computes the effective input–output transformation of the ViT for each sample, and filters out operator-variation term. The resulting attribution operators are inverse-transformed, averaged, and applied element-wise to the input to produce the final DAVE attribution map.}
    \label{fig:implementation}
\end{figure*}

\vspace{-5mm}
\section{Related works}

Model explainability in vision tasks has long focused on post-hoc analysis of convolutional architectures.
However, the prominence of Vision Transformers has required a shift towards addressing the unique transformer-specific artifacts and long-range dependencies. In parallel, approaches that modify the model's architecture have been recently introduced with B-cos networks~\cite{bohle2022b} as an example. In this section, we briefly review these directions and position DAVE among them.

\paragraph{Architecture-agnostic Attribution Methods.}~\looseness=-1Early feature attribution methods mostly rely on gradient information or input sensitivity to identify salient regions. Saliency Maps~\cite{simonyan2013deep} visualize the raw gradient of the output with respect to the input pixels to highlight influential regions. SmoothGrad~\cite{smilkov2017smoothgrad} reduces visual noise by averaging gradients over multiple noisy copies of the input. Deconvolution~\cite{mahendran2016salient} computes attributions by backpropagating through a network while overriding ReLU gradients to propagate only non-negative signals. Guided Backpropagation~\cite{rebuffi2020there} further refines this by combining deconvolution with standard backpropagation to visualize features that specifically activate high-level neurons. Integrated Gradients (IG)~\cite{sundararajan2017axiomatic} provides an axiomatic foundation by integrating local gradients along a straight-line path from a baseline to the input, satisfying properties such as completeness and sensitivity. Grad-CAM~\cite{selvaraju2017grad} uses the gradients of a target class flowing into the feature maps in a chosen layer to produce a localization map highlighting important regions. FullGrad~\cite{srinivas2019full} captures a more comprehensive signal by aggregating gradients with respect to both inputs and model biases. OMENN~\cite{wrualbel2024omenn} proposes dynamic linearity as a way to decompose neural network operations in an input-dependet way to identify a single matrix to explain a model's decisions.

Most of these methods are not specific to ViTs and can be prone to artifacts induced by patch tokenization and attention routing. In contrast, DAVE is designed for ViTs. 

\paragraph{ViT-Specific Methods.}
The shift toward self-attention mechanisms has motivated specialized techniques to handle patch-based interactions and attention artifacts~\cite{lu2025artifacts}. Raw attention visualizations use the model's inherent attention weights to show token interactions, though they often include significant background noise and are not necessarily class-specific. Rollout~\cite{abnar2020quantifying} traces the flow of importance through the entire network by combining attention maps across all layers via matrix multiplication. Attention Flow~\cite{abnar2020quantifying} casts the attention mechanism as a max-flow problem to trace information flow. CheferCAM~\cite{chefer2021transformer} weights self-attention maps by their gradients and aggregates them through the network to provide class-specific explanations. Iterated Integrated Attributions (IIA)~\cite{barkan2023visual} generates explanation maps by iteratively integrating gradients and attention maps across all layers. AttnLRP~\cite{achtibat2024attnlrp} introduces redistribution rules to handle the non-linear softmax operations within the self-attention block. LeGrad~\cite{bousselham2025legrad} utilizes a streamlined approach by summing layer-wise gradients with respect to attention maps, effectively filtering out ``register token" artifacts and capturing hierarchical feature formation.

In contrast, DAVE decomposes the input gradient to isolate a locally equivariant and comparatively stable signal, reducing grid-like artifacts in ViT attributions.

\paragraph{Inherently interpretable Architectures.}
Instead of post-hoc explanations, some approaches modify the model's architecture to be inherently interpretable~\cite{protopnets, bagnets}. B-cos networks~\cite{bohle2022b} enforce stronger alignment between inputs and weights by replacing standard linear layers with B-cos transformations, making the weights themselves a faithful explanation. The B-cos Vision Transformer~\cite{bohle2024b} applies this principle to ViTs, replacing linear transformations within the transformer to promote human-interpretable weight-input alignment. B-cosification~\cite{arya2024b} is a novel technique that fine-tunes existing pre-trained models to adopt these transparent B-cos transformations at a fraction of the cost of training from scratch. 

Unlike these approaches, DAVE improves explanations for standard pretrained ViTs, and we demonstrate that it can also be applied to inherently interpretable models such as B-cos networks.

\section{DAVE}
DAVE interprets attribution as the stable and locally equivariant \emph{effective transformation} that a Vision Transformer applies to its input (\cref{fig:implementation}). Under architectural assumptions of ViTs, it extracts this transformation from the input gradient by decomposing it into a direct input–output transformation and a local variation term (Section~\ref{sec:effective_transform}). The method discards the local variation term, which captures input-dependent sensitivity of internal model mechanisms, and aggregates the remaining transformation over a distribution of inputs. This distribution is constructed to suppress locally non-equivariant components of the transformation (Section~\ref{sec:equivariant_transform}), as well as high-frequency operator fluctuations (Section~\ref{sec:low_pass}), while preserving consistent input-dependent attribution structure.

\subsection{Extracting the effective transformation}\label{sec:effective_transform}
The input gradient is commonly used as a measure of model sensitivity. However, the architectural structure of ViT layers reveals that their gradient entangles two distinct components: the \emph{effective transformation} applied to the input and the variation of this transformation with respect to the input. In this section, we show how these contributions can be decomposed, enabling us to recover the effective transformation used as a baseline attribution signal.

\subsubsection{Layer structure}
We begin by unifying the structure of ViT layers under a single definition, which allows for a principled analysis of their derivatives.

\paragraph{Layer definition.}
Let $V_{\mathrm{in}}$ and $V_{\mathrm{out}}$ be finite-dimensional real vector spaces.
We model each ViT layer $F: V_{\mathrm{in}} \to V_{\mathrm{out}}$ as a differentiable operator-valued map $L : V_{\mathrm{in}} \to \mathbb{L}(V_{\mathrm{in}}, V_{\mathrm{out}})$ followed by a constant bias addition\footnote{$\mathbb{L}(V_{\mathrm{in}}, V_{\mathrm{out}})$ denotes the space of
linear maps from $V_{\mathrm{in}}$ to $V_{\mathrm{out}}$. The map $L$ assigns to
each input $\boldsymbol{X}$ a linear operator $L(\boldsymbol{X})$, yielding an
input-dependent linear transformation, also referred as \emph{dynamic linear}~\cite{bohle2022b,bohle2024b}.}
\begin{equation}\label{eq:vit_layer}
F(\boldsymbol{X})
\;:=\;
L(\boldsymbol{X})(\boldsymbol{X})
\;+\;
\boldsymbol{B}
\end{equation}
where $\boldsymbol{B} \in V_{out}$ represents a constant bias parameter.

\paragraph{Layer realisation.}
Motivated by the architectural structure of ViT layers, we identify layer inputs and
outputs with matrices in $\mathbb{R}^{t \times d_{\mathrm{in}}}$ and $\mathbb{R}^{t \times d_{\mathrm{out}}}$, where $t$ denotes the number of tokens and $d_{\mathrm{in}}, d_{\mathrm{out}}$ token dimensionalities.
Under this identification, we consider two classes of maps $L$:
\begin{equation}\label{eq:layer_representation}
\begin{split}
(\text{I})& \quad
L(\boldsymbol{X})(\boldsymbol{X})
:= \boldsymbol{W}_t(\boldsymbol{X})\,\boldsymbol{X}\,\boldsymbol{W}_d\\
(\text{II})& \quad
L(\boldsymbol{X})(\boldsymbol{X})
:= \boldsymbol{\Phi}(\boldsymbol{X}) \odot \boldsymbol{X}
\end{split}
\end{equation}
where $\boldsymbol{W}_t(\boldsymbol{X}) \in \mathbb{R}^{t \times t}$ is an
input-dependent token-mixing matrix,
$\boldsymbol{W}_d \in \mathbb{R}^{d_{in} \times d_{out}}$ is a static dimensional projection,
and $\boldsymbol{\Phi}(\boldsymbol{X}) \in \mathbb{R}^{t \times d_{in}}$ is an
input-dependent gating matrix. Although generated nonlinearly, these matrices act linearly on $\boldsymbol{X}$
for a fixed input, via matrix multiplication or elementwise multiplication.

These operators cover all ViT components, including attention ($\text{I}$), normalization ($\text{I}$), and pointwise activations ($\text{II}$); details are provided in the Appendix~\ref{app:layers_representations}.

\paragraph{Layer derivative.}
Under the considered layer definition (Equation~\ref{eq:layer_representation}) the derivative operator of $F$ decomposes into the effective transformation $L(\boldsymbol{X})$ and its variation with respect to the input:
\begin{equation}\label{eq:layer_derivative}
\underbrace{D_{\boldsymbol{X}}F}_{\substack{\text{layer}\\\text{derivative}}}
=
\underbrace{L(\boldsymbol{X})}_{\substack{\text{effective}\\\text{transformation}}}
+
\underbrace{\bigl((D_{\boldsymbol{X}}L(\boldsymbol{X})(\cdot))\,\boldsymbol{X}\bigr)}_{\substack{\text{operator}\\\text{variation}}}
\end{equation}

The effective transformation $L(\boldsymbol{X})$ captures the direct, input-conditioned action of
the layer on its input. In contrast, the variation term corresponds to the
derivative of an input-dependent operator and measures its
changes under small input perturbations.

\subsubsection{Effective transformation}
\begin{figure}[!t]
    \centering
    \includegraphics[width=\linewidth]{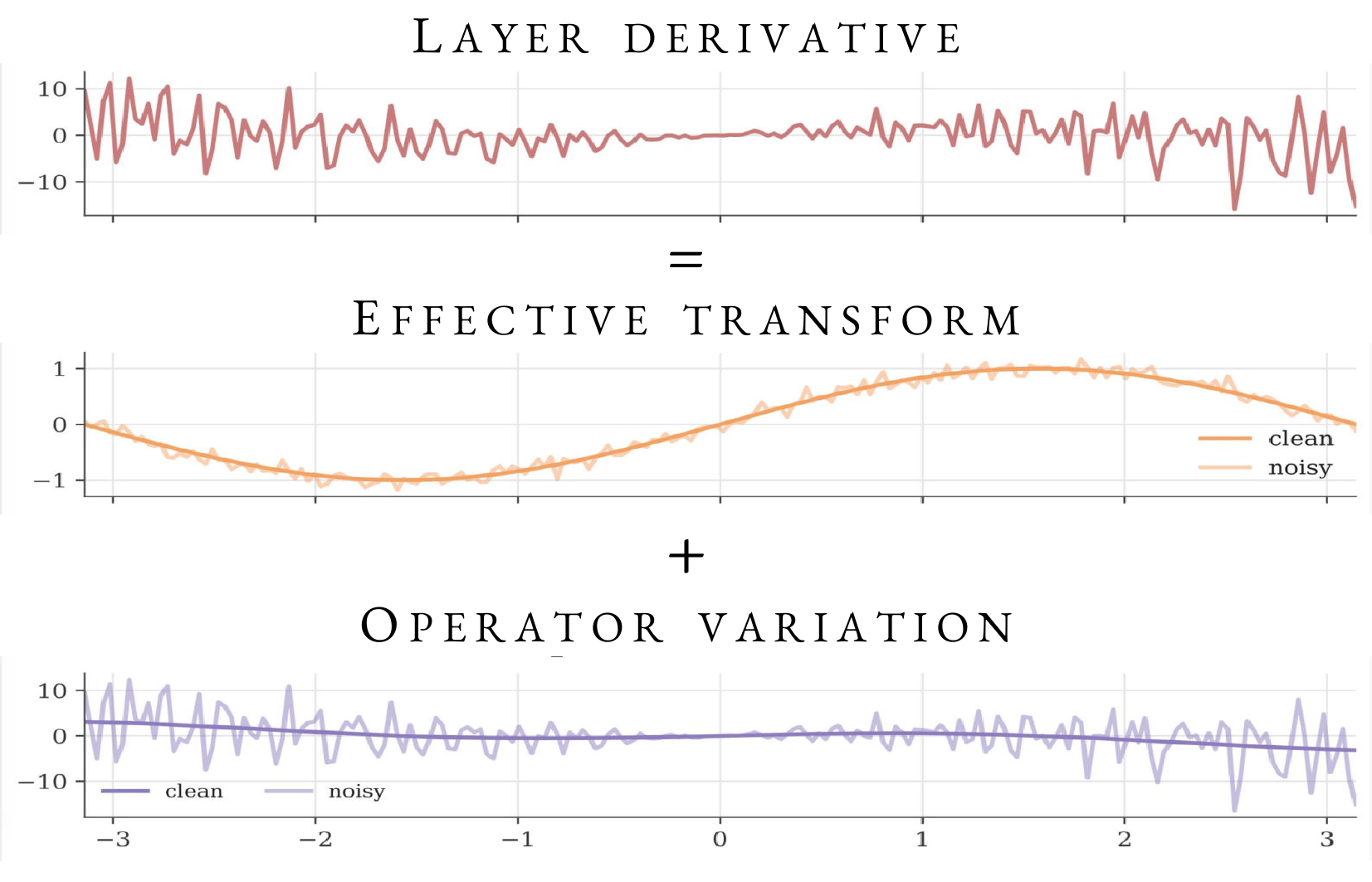}
    \vspace{-5mm}
    \caption{\textbf{Toy example:} illustrating operator variation dominating the layer derivative despite a stable effective transformation (Eq.~\ref{eq:layer_derivative}).
Top: full layer derivative (sum), dominated by operator variation.
Middle: effective transformation with a small perturbation.
Bottom: operator-variation term, where the perturbation is amplified.}
\vspace{-2mm}
    \label{fig:toy}
\end{figure}

Operator variation (Equation~\ref{eq:layer_derivative}) can amplify high-frequency and locally unstable components, leading to gradient-based
attribution signals that are sensitive to small input perturbations
(see Figure~\ref{fig:toy}). 
Since our goal is to explain the effective transformation underlying a specific prediction
rather than infinitesimal sensitivity, we omit the operator-variation term and
retain only the effective transformation $L(\boldsymbol{X})$
(see Figure~\ref{fig:method}, columns~2 and~3).

This yields a pointwise effective operator. Subsequent sections reintroduce stable input-dependent structure by averaging the effective transformation over small neighborhoods of the input, while suppressing high-frequency artifacts.

\paragraph{Effective transformation representation.}
For fixed bases $\mathcal{B}_{\mathrm{in}}$ and $\mathcal{B}_{\mathrm{out}}$ of the
input and output vector spaces of the layer, the effective transformation admits an input-dependent matrix representation
$\boldsymbol{W}_L(\boldsymbol{X}) \in \mathbb{R}^{t d_{\mathrm{out}} \times t d_{\mathrm{in}}}$.
Applied to the input vector $\boldsymbol{x} = [\boldsymbol{X}]_{\mathcal{B}_{\mathrm{in}}}
\in \mathbb{R}^{t d_{\mathrm{in}}}$, this matrix summarizes the direct action of the
layer on its input:

\begin{equation}\label{eq:effective_weight}
[L(\boldsymbol{X})(\boldsymbol{X})]_{\mathcal{B}_{out}} = \boldsymbol{W}_L(\boldsymbol{X})\boldsymbol{x}
\end{equation}

Importantly, $\boldsymbol{W}_L(\boldsymbol{X})$ is not a local sensitivity descriptor (as captured by the full Jacobian), but a matrix representation of the input-conditioned transformation applied to $\boldsymbol{X}$ (see Appendix~\ref{app:effective_weight} for details).

For a network composed of $n$ layers, the effective transformation of the entire model is
obtained by composing the layerwise effective transformations:
\[
\boldsymbol{W}_L(\boldsymbol{X})
=
\prod_{i=n}^1
\boldsymbol{W}_{L_i}(\boldsymbol{X}_{i-1}).
\]
This input-dependent effective weight matrix  combines attention weights, learned projections, normalization statistics, and gating across all layers, and serves as the baseline attribution operator used by DAVE.

\subsection{Extracting an equivariant transformation}\label{sec:equivariant_transform}
\begin{figure*}[!t]
    \centering
    \includegraphics[width=1.0\linewidth]{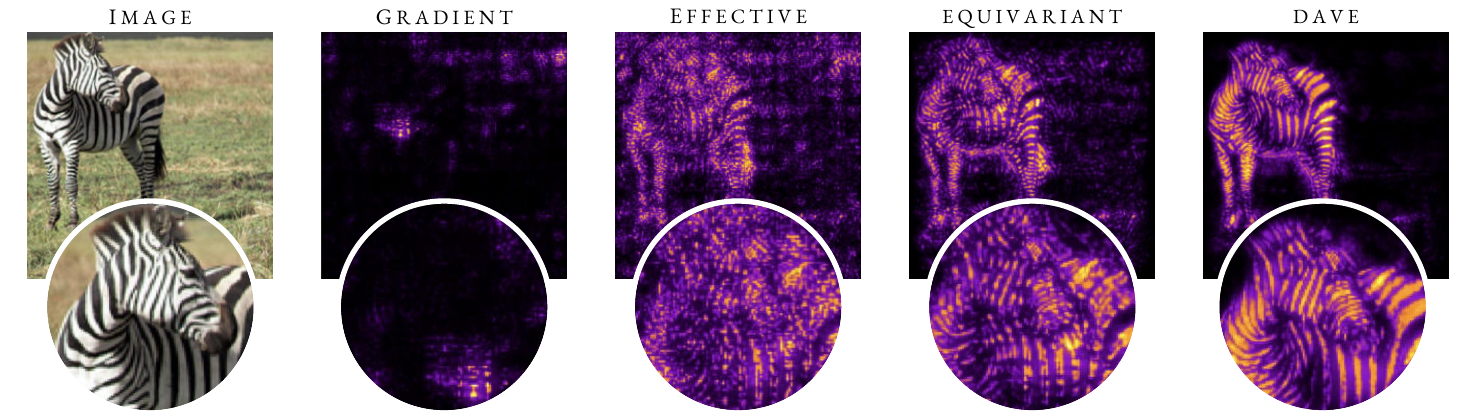}
    \caption{\textbf{Progressive construction of DAVE attribution.} 
    The effective transformation captures the direct input–output action of the network while discarding operator-variation terms.
Equivariant aggregation suppresses architecture-induced artifacts by enforcing local transformation consistency.
DAVE further applies low-pass stabilization, yielding a stable and interpretable attribution map.
Columns show input, input$\times$gradient, input$\times$effective transformation, input$\times$equivariant transformation, and final DAVE attribution, respectively.
}
    \label{fig:method}
\end{figure*}

The effective transformation often exhibits architecture-induced patterns that are stable across inputs, such as grid-like artifacts from patch embedding and attention routing. While these components contribute to the model’s overall behavior, their relative input-invariance limits their usefulness for explaining individual predictions and can produce visually dominant but uninformative attribution patterns (Figure~\ref{fig:method}, column~3). We therefore introduce a Reynolds-inspired~\cite{serre1977linear} operator that suppresses such artifacts by averaging over a local neighborhood of transformed inputs, isolating components that vary consistently under small spatial transformations.

\paragraph{Equivariance criterion.}
Vision models typically recognize the same semantic content under small spatial transformations of the input, resulting in approximately invariant predictions. Since attribution aims to localize the visual evidence supporting a prediction, attribution maps should transform equivariantly under such transformations.

Formally, given an attribution map $\boldsymbol{W}_L(\boldsymbol{X})$, and a group of spatial transformations $\tau: V_{\mathrm{in}} \to V_{\mathrm{in}}$ we aim to
extract a locally equivariant component
satisfying\footnote{
We assume that the network outputs a
scalar logit ($\dim(V_{\mathrm{out}})=1$), so that the effective transformation
$\boldsymbol{W}_L(\boldsymbol{X})$ admits a spatial representation with the same
dimensionality as the input, as in gradient-based attribution methods.
}:
\begin{equation}\label{eq:equivariance}
\boldsymbol{W}^{\mathrm{eq}}_L(\boldsymbol{X})
\;\approx\;
\bigl[\tau^{-1} \circ \boldsymbol{W}_L \circ \tau\bigr](\boldsymbol{X}),
\end{equation}
for transformations $\tau$ near the identity. This condition enforces that the attribution map transforms consistently under small spatial transformations of the input, providing a criterion for isolating locally equivariant attribution components.

\paragraph{Equivariant transformation.}
To suppress components of $\boldsymbol{W}_L(\boldsymbol{X})$ that violate this
local consistency, we apply a Reynolds-inspired filtering operator
defined over a group of spatial transformations.

Let $G$ be a compact group of spatial
transformations (e.g., rotations), and let $\nu$ be a probability measure
supported in a neighborhood of the identity, enforcing locality of the equivariance constraint. We define:
\begin{equation}\label{eq:averaging}
\boldsymbol{W}^{eq}_L(\boldsymbol{X})
=
\int_G
[\tau^{-1}\circ\boldsymbol{W}_L\circ\tau](\boldsymbol{X})
\, d\nu(\tau)
\end{equation}

This operator suppresses components of $\boldsymbol{W}_L(\boldsymbol{X})$ that
violate local equivariance, while retaining attribution signals that transform
consistently under transformations in $G$ (see Appendix~\ref{app:equivariant} for details).

\paragraph{Controlled equivariant variation.}
While discarding the operator-variation term removes infinitesimal sensitivity of the transformation, the subsequent Reynolds-style averaging captures transformation-consistent changes of the effective operator across a local neighborhood of inputs, corresponding to controlled, equivariant variation, while avoiding amplification of high-frequency instabilities (see Figure~\ref{fig:method} columns 3 and 4 for visual comparison).


\subsection{Low-pass filtering}\label{sec:low_pass}
To suppress high-frequency artifacts in the attribution signal, we apply
a low-pass filtering operation.
Specifically, we perform local averaging of the equivariant effective
transformation by evaluating it under small input perturbations drawn from a local probability distribution and taking the expectation. For Gaussian
perturbations, this operation is equivalent to convolution 
with the corresponding smoothing kernel $\mathcal{K}$, as shown in
Appendix~\ref{app:stochastic_convolution}:
\begin{equation}
\mathbb{E}_{\boldsymbol{\epsilon} \sim \mathcal{N}(0,\Sigma)}
\left[
\boldsymbol{W}^{\mathrm{eq}}_L(\boldsymbol{X} + \boldsymbol{\epsilon})
\right]
=
(\boldsymbol{W}^{\mathrm{eq}}_L \ast \mathcal{K})(\boldsymbol{X})
\end{equation}

This convolution attenuates attribution components that are unstable under small input variations (see Figure~\ref{fig:method} column 5). Although this
operation is analogous in form to SmoothGrad-style~\cite{smilkov2017smoothgrad} averaging, it is applied to
the equivariant effective transformation rather than to raw input gradients,
thereby stabilizing the attribution without reintroducing operator-variation artifacts.

\subsection{DAVE framework}\label{sec:dave_full}
We summarize the DAVE attribution framework by combining equivariant filtering
and low-pass stabilization into a single expression. The final
distribution-aware effective transformation (i.e., the attribution-relevant
effective input–output transformation) is obtained by taking the expectation of
the effective transformation under local spatial transformations and small input
perturbations. For a Gaussian smoothing kernel, this yields:
\begin{equation}\label{eq:final_operator}
\tilde{\boldsymbol{W}}^{\mathrm{eq}}_L(\boldsymbol{X})
=
\mathbb{E}_{\tau \sim \nu,\;\boldsymbol{\epsilon} \sim \mathcal{N}(0,\Sigma)}
\left[
\tau^{-1}\!\left(
\boldsymbol{W}_L(\tau(\boldsymbol{X} + \boldsymbol{\epsilon}))
\right)
\right],
\end{equation}

where $\tau \sim \nu$ denotes a spatial transformation drawn from a probability
measure supported in a neighborhood of the identity, and
$\boldsymbol{\epsilon} \sim \mathcal{N}(0,\Sigma)$ denotes Gaussian noise.

\paragraph{Attribution map.}
The DAVE attribution map is obtained by applying this
distribution-aware effective transformation to the input via element-wise
multiplication:
\begin{equation}
\mathrm{DAVE}(\boldsymbol{X})
=
\tilde{\boldsymbol{W}}^{\mathrm{eq}}_L(\boldsymbol{X}) \odot \boldsymbol{X}
\end{equation}

This element-wise product follows from the matrix form of the effective transformation (Equation~\ref{eq:effective_weight}): the filtered operator defines an input-dependent linear map, so each input dimension contributes proportionally to its value weighted by the corresponding entry of the effective transformation.

\paragraph{Practical realisation.}
We compute the effective transformation by formulating network layers according to Equation~\ref{eq:layer_representation} and detaching the input dependence of the operators during the forward pass. The layerwise effective transformations are then aggregated during backpropagation, yielding the effective transformation of the full model. This procedure is closely related to implementation techniques used in \cite{bohle2022b,bohle2024b,wrualbel2024omenn} and allows the effective transformation to be computed with a single modified forward and backward pass.

The expectation in Equation~\ref{eq:final_operator} is approximated via Monte Carlo sampling. We draw a small number of spatial transformations from the distribution $\nu$ and evaluate the effective transformation at the corresponding transformed inputs, additionally applying a small Gaussian perturbation. The final distribution-aware effective transformation is obtained by averaging the inverse-transformed attribution operators across samples (Figure~\ref{fig:implementation}).


\begin{figure*}[!t]
  \centering
  \includegraphics[width=0.465\textwidth]{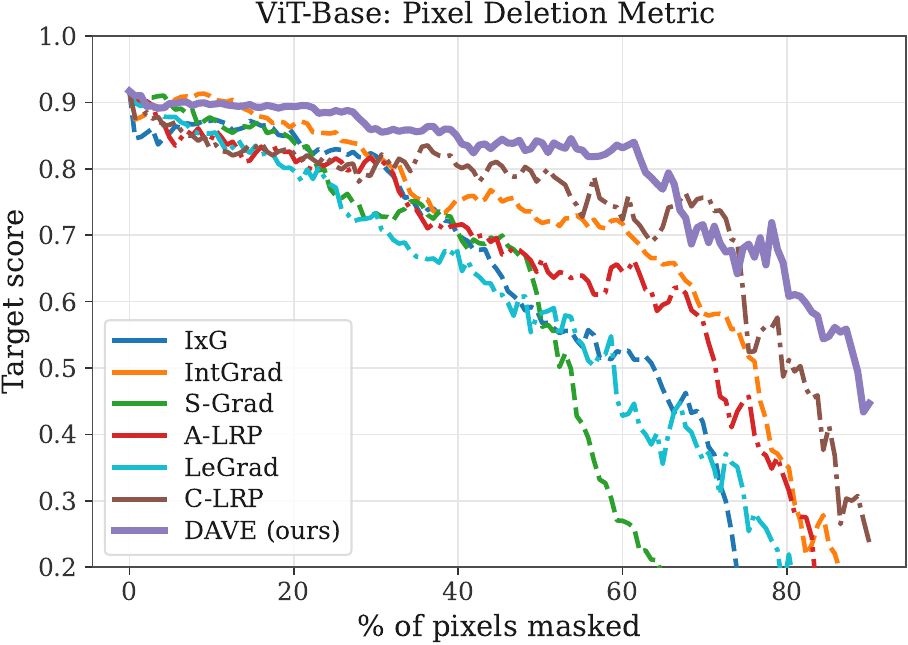}\hfill
  \includegraphics[width=0.451\textwidth]{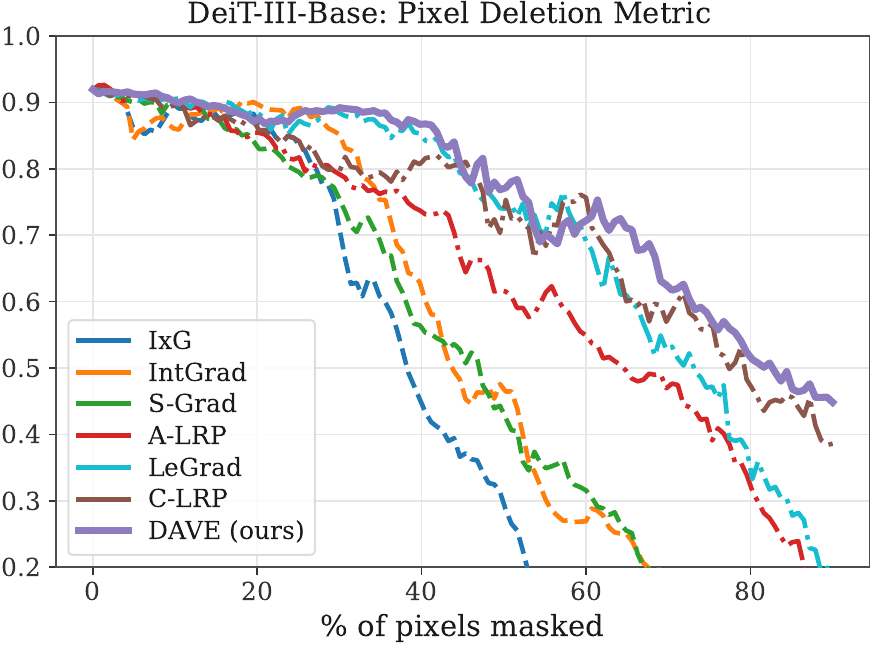}
  \vspace{-1mm}
  \caption{\textbf{Pixel deletion faithfulness.} Target-class probability versus fraction of pixels removed on ViT-B/16 (left) and DeiT-III-B/16 (right). DAVE (\textcolor{davemauve}{purple}) yields the flattest curves (higher AUC), indicating the strongest stability under deletion.}
  \label{fig:pixel_del_side_by_side}
\end{figure*}


\section{Experimental Setup}
\label{sec:exp_setup}

In this section, we describe the evaluation setup for \textbf{DAVE}, our distribution-aware attribution method for Vision Transformers. Code is provided in supplementary material.

\paragraph{Models.} We evaluate DAVE on standard ViT-based classifiers, including ViT-B/16~\cite{dosovitskiy2020image}, DeiT-B/16, and DeiT-III-B/16~\cite{touvron2021training,touvron2022deit}, as well as a self-supervised DINO ViT-B/16 model~\cite{caron2021emerging}, all at $224{\times}224$ input resolution. For all models, we use the authors' publicly released pretrained checkpoints from their official repositories. We additionally evaluate on inherently interpretable \bcos models (B-cos-ViT and B-cos-ViT-C)~\cite{bohle2024b}, which provide class-specific explanations and serve as strong transparent baselines.


\paragraph{Datasets.}
~\looseness=-1We use the publicly available ImageNet-1k~\cite{imagenet} dataset for all attribution evaluations and qualitative comparisons. For localization experiments, we evaluate on the ImageNet validation set using the official bounding-box annotations (metrics defined below).

\begin{table}[!t]
\centering
\tiny
\caption{\textbf{Localization on conventional models.} GridPG (left) and EnergyPG (right) on ImageNet-1k. $\Delta$ denotes the improvement of DAVE over the best competing method (higher is better).}
\setlength{\tabcolsep}{2.0pt}
\renewcommand{\arraystretch}{1.08}

\resizebox{\columnwidth}{!}{%
\begin{tabular}{lcccccccc}
\toprule
& \multicolumn{4}{c}{\textbf{GridPG} (\%)} & \multicolumn{4}{c}{\textbf{EnergyPG} (\%)} \\
\cmidrule(lr){2-5}\cmidrule(lr){6-9}
Method & ViT-B & DeiT-B & D-III-B & DINO-B & ViT-B & DeiT-B & D-III-B & DINO-B \\
\midrule
I$\times$G   & 32.67 & 30.25 & 30.01 & 33.28 & 55.33 & 62.72 & 67.32 & 69.98 \\
IntGrad      & 39.86 & 36.11 & 31.68 & 36.98 & 58.51 & 64.22 & 68.12 & 72.15 \\
S-Grad       & 34.27 & 30.18 & 31.48 & 33.13 & 56.02 & 60.78 & 68.10 & 70.93 \\
LeGrad       & 47.71 & 42.58 & 34.62 & 28.96 & 80.06 & 77.83 & 77.54 & 82.26 \\
A-LRP        & 58.40 & 54.63 & 53.84 & 37.49 & 60.75 & 68.16 & 77.65 & 75.98 \\
C-LRP        & 54.98 & 55.47 & 52.27 & 49.99 & \textbf{80.82} & 79.62 & 81.94 & 81.56 \\
\rowcolor{davemauve!25}
DAVE (ours)  & \textbf{60.19} & \textbf{63.52} & \textbf{65.76} & \textbf{51.33} & 78.60 & \textbf{82.23} & \textbf{82.43} & \textbf{83.38} \\
\midrule
$\Delta$ 
& \textbf{\inc{+1.79}} & \textbf{\inc{+8.05}} & \textbf{\inc{+11.92}} & \textbf{\inc{+1.35}}
& \textcolor{red}{-2.22} & \textbf{\inc{+2.61}} & \textbf{\inc{+0.49}} & \textbf{\inc{+1.12}} \\
\bottomrule
\end{tabular}%
}
\label{tab:pg_vit_deit_pct}
\end{table}

\begin{table}[!t]
\centering
\caption{\textbf{Localization on \bcos models.} GridPG (left) and EnergyPG (right) on ImageNet-1k. $\Delta$ denotes the improvement of DAVE over the inherent \bcos explanation (higher is better).}
\scriptsize
\setlength{\tabcolsep}{3pt}
\renewcommand{\arraystretch}{1.12}
\begin{tabularx}{\columnwidth}{l *{2}{>{\centering\arraybackslash}X}  *{2}{>{\centering\arraybackslash}X}}
\toprule
& \multicolumn{2}{c}{\textbf{Grid PG} (\%)} & \multicolumn{2}{c}{\textbf{Energy PG} (\%)} \\
\cmidrule(lr){2-3}\cmidrule(lr){4-5}
Method & B-cos-ViT & B-cos-ViT-C & B-cos-ViT & B-cos-ViT-C \\
\midrule
I$\times$G         & 55.71 & 55.38 & 65.29 & 66.72 \\
IntGrad          & 57.98 & 57.68 & 65.33 & 67.75 \\
S-Grad  & 60.27 & 61.30 & 68.16 & 69.59 \\
\bcos        & 79.67 & 87.66 & 69.41 & 75.61 \\
%
\rowcolor{davemauve!25}
DAVE (ours)       & \textbf{84.00} & \textbf{88.43} & \textbf{78.55} & \textbf{79.63} \\
\midrule
$\Delta$  & \textbf{\inc{+4.33}} & \textbf{\inc{+0.77}} & \textbf{\inc{+9.14}} & \textbf{\inc{+4.02}} \\
\bottomrule
\end{tabularx}

\vspace{-6mm}
\label{tab:pg_bcos_pct}
\end{table}

\paragraph{Baselines.}
We compare against representative gradient-based and transformer-specific attribution methods:
Input$\times$Gradient (I$\times$G)~\cite{shrikumar2017learning}, Integrated Gradients (IntGrad)~\cite{sundararajan2017axiomatic}, SmoothGrad (S-Grad)~\cite{smilkov2017smoothgrad}, LeGrad~\cite{bousselham2025legrad}, AttnLRP (A-LRP)~\cite{achtibat2024attnlrp}, and Chefer-LRP (C-LRP)~\cite{chefer2021transformer}. For B-cos networks, we additionally include the model inherent B-cos explanations~\cite{bohle2024b}.

\paragraph{Attribution map generation.}
All methods produce class-conditional attribution maps for the predicted class at the model input resolution. For methods that operate at patch resolution (\textit{e.g.,} C-LRP, LeGrad), we upsample attribution maps to pixel space using bilinear interpolation for evaluation. Unless stated otherwise, we follow the authors' official implementations and default hyperparameters.

We evaluate explanations along two axes: \textbf{localization} and \textbf{faithfulness}.

\paragraph{Localization metrics.}
We quantify attribution localization using two pointing-game metrics:  \textit{Grid Pointing Game (GridPG)} and \textit{Energy Pointing Game (EnergyPG)}.  GridPG measures whether the maximally attributed location falls inside the ground-truth object region, while EnergyPG measures the fraction of attribution mass contained inside the ground-truth bounding box (higher is better for both).

\myparagraph{Faithfulness metric.}
We evaluate faithfulness using \textit{pixel deletion} curves for all models. We progressively remove pixels in order of increasing attribution (least to most important) and track the target-class probability. We plot the target-class probability versus the fraction of pixels removed; flatter curves highlight more stable attributions that are consistent with the model's decision. We follow standard deletion protocols from prior
work~\cite{chefer2021transformer}. 

Additional evaluation details are in the Appendix~\ref{app:impl_details}.

\begin{figure}[!h]
  \centering
  \includegraphics[width=1.0\columnwidth]{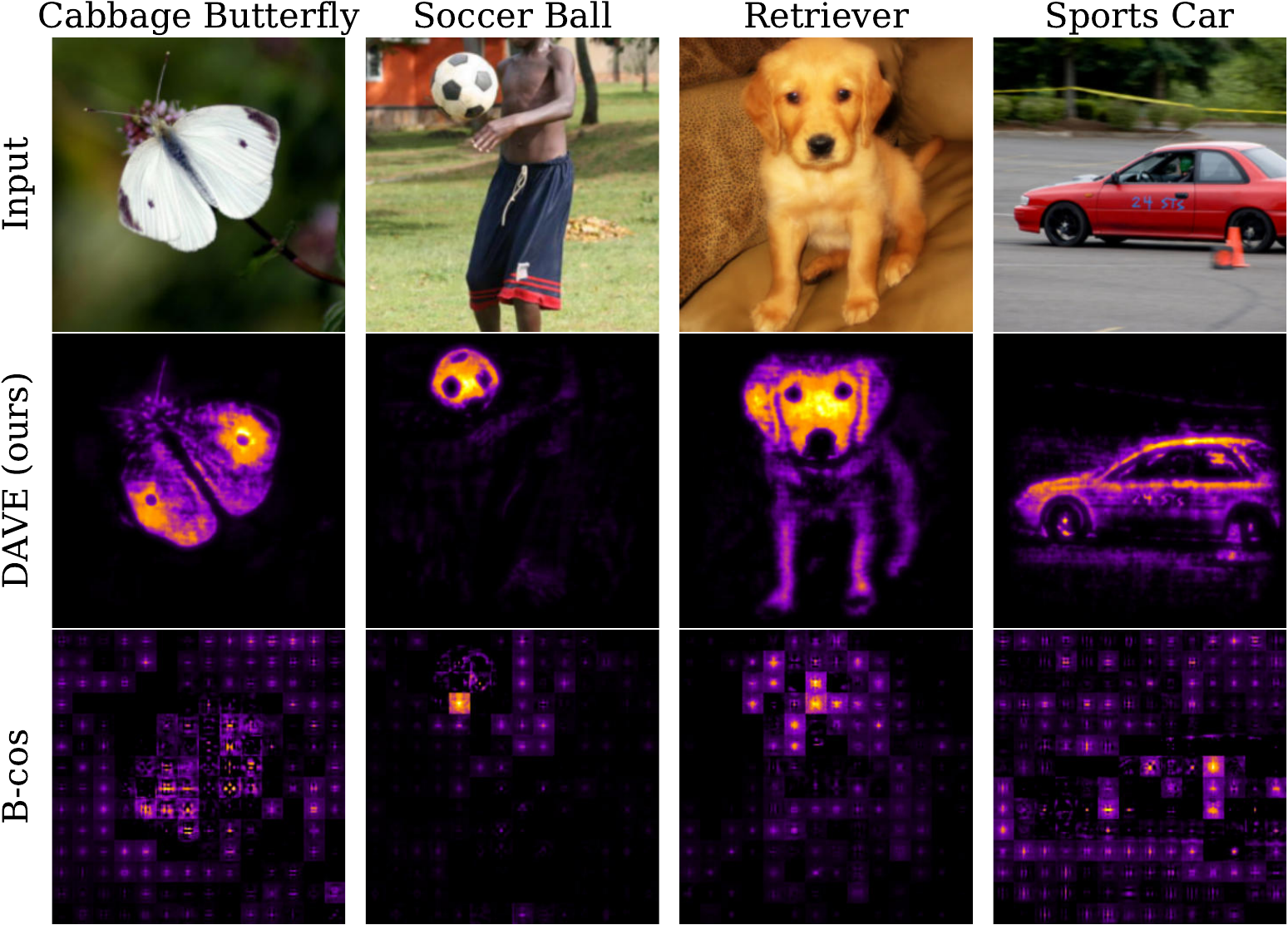}
\caption{\textbf{Improved attributions for B-cos models.} We show ImageNet inputs (top), DAVE (middle), and inherent B-cos explanations (bottom). DAVE produces sharper, more object-aligned attributions with reduced background responses.
}

\vspace{-3mm}
  \label{fig:dave_vs_bcos_main}
\end{figure}
\section{Results}
\label{sec:results}

In this section, present quantitative and qualitative results for DAVE compared to prior attribution methods.

\subsection{Main Results}

\paragraph{Localization.}


Table~\ref{tab:pg_vit_deit_pct} reports localization (\%) on conventional ViTs. DAVE improves \textbf{GridPG} across all evaluated models. On ViT-B/16, DAVE achieves \textbf{60.19}\% GridPG, improving over the strongest competing method by \textbf{+1.79} p.p. On DeiT-B/16 and DeiT-III-B/16, DAVE yields \textbf{63.52}\% and \textbf{65.76}\% GridPG, with gains of \textbf{+8.05} and \textbf{+11.92} p.p., respectively. On the self-supervised DINO-B/16 backbone, DAVE reaches \textbf{51.33}\% GridPG, improving over the strongest competing method by \textbf{+1.35} p.p. For \textbf{EnergyPG}, DAVE achieves the best results on DeiT-B/16 (\textbf{82.23}\%) and DeiT-III-B/16 (\textbf{82.43}\%), and is competitive on ViT-B/16. On DINO-B/16, DAVE achieves \textbf{83.38}\% EnergyPG, improving over the strongest competing method by \textbf{+1.12} p.p.

Table~\ref{tab:pg_bcos_pct} reports localization (\%) on \bcos models. DAVE improves over the inherent \bcos explanations on both GridPG and EnergyPG. For B-cos-ViT, DAVE achieves \textbf{84.00}\% GridPG and \textbf{78.55}\% EnergyPG, improving over \bcos by \textbf{+4.33} and \textbf{+9.14} p.p., respectively. For B-cos-ViT-C (with a convolutional stem), DAVE achieves \textbf{88.43}\% GridPG and \textbf{79.63}\% EnergyPG, corresponding to gains of \textbf{+0.77} and \textbf{+4.02} p.p.\ over \bcos.


\paragraph{Faithfulness via pixel deletion.}
Figure~\ref{fig:pixel_del_side_by_side} shows pixel deletion curves on ViT-B/16 (left) and DeiT-III-B/16 (right), tracking target-class probability versus the fraction of pixels removed. DAVE remains the most stable under deletion, exhibiting the flattest curves compared to prior methods.

\begin{figure*}[!t]
  \centering
  \includegraphics[width=1.0\textwidth]{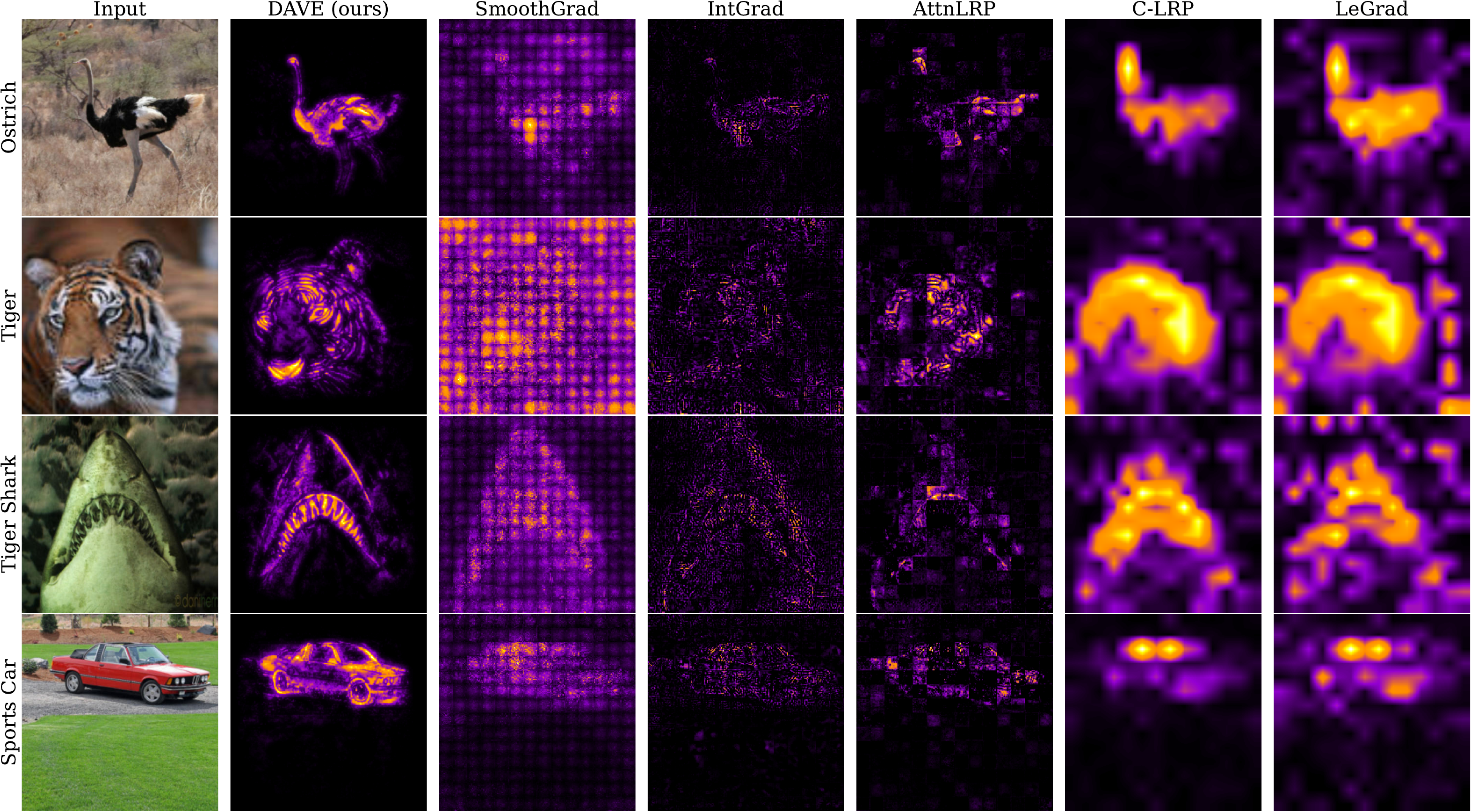}
\caption{\textbf{Qualitative comparison of attribution maps.} Dave yields sharper, more object-aligned explanations with reduced background noise compared to prior methods. Note that for a set of ImageNet-1k examples (rows), we show the input image (left) and attribution maps produced by \textbf{DAVE (ours)}, SmoothGrad, IntGrad, AttnLRP, C-LRP, and LeGrad (columns). 
}
\vspace{-3mm}
  \label{fig:dave_vs_rest_deit3_base_4}
\end{figure*}

\paragraph{Qualitative comparisons.}
Figures~\ref{fig:dave_vs_rest_deit3_base_4} and~\ref{fig:dave_vs_bcos_main} provide qualitative comparisons on ImageNet-1k validation examples. In \cref{fig:dave_vs_rest_deit3_base_4}, several prior methods exhibit patch-grid artifacts and more diffuse explanations, particularly SmoothGrad (col.~3), IntGrad (col.~4), and transformer-specific baselines (cols.~5--6). In contrast, DAVE (col.~2) produces sharper, more object-aligned attribution maps with reduced noise, consistent with the localization gains in Tables~\ref{tab:pg_vit_deit_pct}, and ~\ref{tab:pg_bcos_pct}. 

For \bcos ViTs (\cref{fig:dave_vs_bcos_main}) trained without a convolutional stem, the inherent \bcos explanations are comparatively noisy, whereas DAVE yields more fine-grained and object-specific attributions.

For additional qualitative results on additional models, class-consistency and varied classes see Appendix~\ref{app:qualitative}.



\subsection{Ablation Studies and Analysis}

\paragraph{Computational Complexity.}
 DAVE attribution has linear complexity in the number of neighborhood samples, i.e.\ $\mathcal{O}(\text{Samples})$, as described in Section~\ref{sec:dave_full}. A comparison with other attribution methods is provided in Table~\ref{tab:attribution_complexity}.

\paragraph{Convergence w.r.t.\ Number of Samples.}
All experiments use 50 samples, after which we observe no further improvement in
evaluation metrics across models; attribution convergence analysis is provided
in Appendix~\ref{app:ablation_analysis}.

\paragraph{Rotation Sensitivity.}
~\looseness=-1We define a stable attribution neighborhood using rotations within $(-20^\circ, 20^\circ)$. This choice is motivated by the observed low model sensitivity for small rotations, measured as the absolute change in softmax probabilities (see Appendix~\ref{app:ablation_analysis}).
Additional analysis of transformation ranges, input perturbations, and setup choices is provided in the Appendix~\ref{app:setup}.








\begin{table}[t]
\centering
\setlength{\tabcolsep}{2.0pt}
\renewcommand{\arraystretch}{0.88}
\caption{\textbf{Computational and memory complexity of attribution methods relative to one forward pass.} For DAVE, $\text{Samples}$ refers to a number of neighborhood samples.}
\label{tab:attribution_complexity}
\begin{tabularx}{\columnwidth}{l *{2}{>{\centering\arraybackslash}X}}
\toprule
\textbf{Method} & \textbf{Computation} & \textbf{Memory} \\
\midrule
I$\times$G   & $\mathcal{O}(1)$              & $\mathcal{O}(\sqrt{\mathrm{Layers}})$ \\
IntGrad      & $\mathcal{O}(\mathrm{Steps})$ & $\mathcal{O}(\sqrt{\mathrm{Layers}})$ \\
S-Grad       & $\mathcal{O}(\mathrm{Steps})$ & $\mathcal{O}(\sqrt{\mathrm{Layers}})$ \\
LeGrad       & $\mathcal{O}(\mathrm{Steps})$ & $\mathcal{O}(\sqrt{\mathrm{Layers}})$ \\
C-LRP         & $\mathcal{O}(1)$              & $\mathcal{O}(\sqrt{\mathrm{Layers}})$ \\
A-LRP        & $\mathcal{O}(\mathrm{Steps})$ & $\mathcal{O}(\sqrt{\mathrm{Layers}})$ \\
\rowcolor{davemauve!25}
DAVE (ours)  & $\mathcal{O}(\mathrm{Samples})$ & $\mathcal{O}(\sqrt{\mathrm{Layers}})$ \\
\bottomrule
\end{tabularx}
\vspace{-2mm}
\end{table}

\section{Conclusions}

In this work, we introduced DAVE, a mathematically grounded framework designed to overcome the challenges of stability and resolution in ViTs explainability. By performing a structured decomposition of the input gradient, DAVE successfully isolates the effective transformation of the model from architecture-induced artifacts and input-dependent noise. Experimental results across multiple benchmarks demonstrate that DAVE consistently outperforms existing state-of-the-art methods in producing spatially precise and class-consistent attributions. Furthermore, we showed that DAVE is architecture-versatile, providing robust explanations for both standard ViTs and inherently interpretable \bcos models. It is also robust across pretraining paradigms--fully-supervised and self-supervised.

Despite these strengths, the use of an averaging operator to isolate stable components introduces higher computational overhead compared to raw gradient methods. Future work could reduce this overhead via more efficient sampling. Additionally, automatically selecting the transformation group without manual tuning would better adapt across models.



\paragraph{Impact Statement}

This research contributes to the broader field of Explainable AI (XAI) by providing a more reliable mechanism for auditing high-capacity vision models. As ViTs are increasingly integrated into high-stakes applications such as medical diagnostics and autonomous driving, the ability to verify that a model is attending to relevant visual features rather than architectural artifacts is critical for safety and trust.

By producing stable, high-resolution attribution maps, DAVE can assist practitioners in (1) Identifying Model Bias (2) Model Debugging by providing visual evidence to understand why a model might misclassify specific samples. (3) Enhancing Human-AI Collaboration by offering explainable signals that align more closely with human visual perception, thereby facilitating better decision-making in collaborative environments.

Lastly, this work supports the development of more transparent and accountable machine learning systems, reducing the opaque nature of state-of-the-art computer vision models.

\section*{Acknowledgments}

The work was funded by the "Interpretable and Interactive Multimodal Retrieval in Drug Discovery" project. The "Interpretable and Interactive Multimodal Retrieval in Drug Discovery" project (FENG.02.02-IP.05-0040/23) is carried out within the First Team programme of the Foundation for Polish Science co-financed by the European Union under the European Funds for Smart Economy 2021-2027 (FENG). 

Some experiments were performed on servers purchased with funds from the flagship project entitled ``Artificial Intelligence Computing Center Core Facility'' from the DigiWorld Priority Research Area within the Excellence Initiative -- Research University program at Jagiellonian University in Kraków.

We gratefully acknowledge Polish high-performance computing infrastructure PLGrid (HPC Center: ACK Cyfronet AGH) for providing computer facilities and support within computational grant no. PLG/2025/018158


\bibliography{example_paper}
\bibliographystyle{icml2026}

\clearpage
\onecolumn
\twocolumn
\appendix
\newpage
\newpage



\section{Mathematical derivations}
\subsection{Layers representations}
\label{app:layers_representations}

In this section, we represent Vision Transformer layers according to the unified layer formulation introduced in
Section~\ref{sec:effective_transform}. We adopt the layer definition
(Equation~\ref{eq:vit_layer}) together with the two operator forms
(Equation~\ref{eq:layer_representation}), and show how we represent principal components
of Vision Transformer (ViT) architectures within this framework.

Specifically, we consider fully-connected layers, self-attention
and multi-head self-attention, layer normalization, pointwise nonlinearities
(GELU and Swish), and residual connections.

In all derivations, the input to a layer is denoted by
\(\boldsymbol{X} \in \mathbb{R}^{t \times d_{\mathrm{in}}}\), representing a
sequence of length \(t\) with \(d_{\mathrm{in}}\)-dimensional token embeddings.

\subsubsection{Fully-Connected}
A fully-connected layer acting independently on each token can be represented within the operator form~(I) of Equation~\ref{eq:layer_representation} by defining the operator-valued map
\[
L(\boldsymbol{X})(\boldsymbol{X}) = \boldsymbol{I}_t\,\boldsymbol{X}\,\boldsymbol{W}_d,
\]
where \(\boldsymbol{I}_t \in \mathbb{R}^{t \times t}\) is the identity matrix over tokens and
\(\boldsymbol{W}_d \in \mathbb{R}^{d_{\mathrm{in}} \times d_{\mathrm{out}}}\) is the learned feature projection.
Substituting this expression into the layer definition (Equation~\ref{eq:vit_layer}) yields the standard fully-connected transformation
\[
F(\boldsymbol{X}) 
= \boldsymbol{X}\boldsymbol{W}_d + \boldsymbol{B}.
\]
In this case, the operator \(L(\boldsymbol{X})\) is input-independent, and the corresponding operator-variation term in the layer derivative is identically zero.

\subsubsection{Self-attention}
We consider a single-head self-attention layer.
Let \(\boldsymbol{W}_Q, \boldsymbol{W}_K, \boldsymbol{W}_V \in \mathbb{R}^{d_{\mathrm{in}} \times d_h}\)
denote the query, key, and value projections, and let
\(\boldsymbol{W}_O \in \mathbb{R}^{d_h \times d_{\mathrm{out}}}\) denote the output projection.
Given an input \(\boldsymbol{X} \in \mathbb{R}^{t \times d_{\mathrm{in}}}\), the attention weights are computed as
\[
\boldsymbol{A}(\boldsymbol{X}) =
\mathrm{softmax}\!\left(
\frac{\boldsymbol{X}\boldsymbol{W}_Q (\boldsymbol{X}\boldsymbol{W}_K)^\top}{\sqrt{d_h}}
\right)
\in \mathbb{R}^{t \times t},
\]
with the softmax applied row-wise.

This operation admits a representation of the form~(I) in
Equation~\ref{eq:layer_representation} by defining the operator-valued map
\[
L(\boldsymbol{X})(\boldsymbol{X})
=
\boldsymbol{W}_t(\boldsymbol{X})\,\boldsymbol{X}\,\boldsymbol{W}_d,
\]
where \(\boldsymbol{W}_t(\boldsymbol{X}) = \boldsymbol{A}(\boldsymbol{X})\) is the
input-dependent token-mixing matrix and
\(\boldsymbol{W}_d = \boldsymbol{W}_V \boldsymbol{W}_O\) is the composed feature
projection.
Substituting this operator into the layer definition
(Equation~\ref{eq:vit_layer}) yields the self-attention transformation
\[
F(\boldsymbol{X})
=
\boldsymbol{A}(\boldsymbol{X})\,\boldsymbol{X}\boldsymbol{W}_V\boldsymbol{W}_O
+ \boldsymbol{B},
\]
where \(\boldsymbol{B} \in \mathbb{R}^{t \times d_{\mathrm{out}}}\) is the bias associated
with the output projection.
If the value projection includes an additive bias, it can be interpreted as a
preceding fully-connected layer and absorbed into the formulation accordingly.

\subsubsection{Multi-head self-attention}
Multi-head self-attention extends the single-head formulation by computing
\(H\) attention heads in parallel, each operating on a distinct feature
subspace.
For head \(h \in \{1,\dots,H\}\), let
\(\boldsymbol{A}^{(h)}(\boldsymbol{X}) \in \mathbb{R}^{t \times t}\) denote the
corresponding attention matrix.
Within the operator form~(I) of
Equation~\ref{eq:layer_representation}, the resulting token-mixing operator can
be expressed as
\[
\boldsymbol{W}_t(\boldsymbol{X})
=
\sum_{h=1}^H
\boldsymbol{A}^{(h)}(\boldsymbol{X}) \otimes \boldsymbol{P}_h,
\]
where \(\boldsymbol{P}_h\) denotes the projection onto the feature subspace
associated with head \(h\), and \(\otimes\) denotes the Kronecker product.
The feature projection \(\boldsymbol{W}_d\) accounts for the head-wise value
projections, concatenation of head outputs, and the subsequent output
projection.
Substituting these definitions into the layer formulation
(Equation~\ref{eq:vit_layer}) recovers the standard multi-head self-attention
operation.

\subsubsection{Layer normalization}
We consider Layer Normalization applied independently to each token.
Given an input \(\boldsymbol{X} \in \mathbb{R}^{t \times d_{\mathrm{in}}}\), let
\(\boldsymbol{\mu}(\boldsymbol{X}) \in \mathbb{R}^{t \times 1}\) and
\(\boldsymbol{\sigma}(\boldsymbol{X}) \in \mathbb{R}^{t \times 1}\) denote the
per-token mean and standard deviation across the feature dimension.
Layer normalization acts as
\[
\mathrm{LN}(\boldsymbol{X})
=
\frac{\boldsymbol{X} - \boldsymbol{\mu}(\boldsymbol{X})\mathbf{1}^\top}
{\boldsymbol{\sigma}(\boldsymbol{X})}
\odot \boldsymbol{\gamma}
+ \boldsymbol{\beta},
\]
where \(\boldsymbol{\gamma}, \boldsymbol{\beta} \in \mathbb{R}^{d_{\mathrm{in}}}\)
are learned affine parameters.

We express this operation within the operator form~(I) of Equation~\ref{eq:layer_representation}.
Mean subtraction corresponds to multiplication by a fixed centering matrix
\[
\boldsymbol{W}_d
=
\boldsymbol{I}_{d_{\mathrm{in}}}
-
\frac{1}{d_{\mathrm{in}}}\mathbf{1}\mathbf{1}^\top
\in \mathbb{R}^{d_{\mathrm{in}} \times d_{\mathrm{in}}},
\]
which defines the feature-space component of the operator.
The variance normalization induces an input-dependent, token-wise scaling,
captured by a diagonal matrix
\[
\boldsymbol{W}_t(\boldsymbol{X})
=
\mathrm{diag}\!\left(\boldsymbol{\sigma}(\boldsymbol{X})^{-1}\right)
\in \mathbb{R}^{t \times t}.
\]

Accordingly, we formulate operator-valued map for normalization as:
\[
L(\boldsymbol{X})(\boldsymbol{X})
=
\boldsymbol{W}_t(\boldsymbol{X})\,\boldsymbol{X}\,\boldsymbol{W}_d
\]
The learned affine parameters \((\boldsymbol{\gamma}, \boldsymbol{\beta})\)
are treated as a subsequent fully-connected layer.
Substituting this operator into the layer definition
(Equation~\ref{eq:vit_layer}), and composing with the affine map, recovers the standard Layer Normalization transformation.

\subsubsection{GELU / Swish}
We express pointwise nonlinearities used in Vision Transformers, such as GELU and Swish,
within the gating form~(II) of
Equation~\ref{eq:layer_representation}.
Consider a scalar nonlinearity \(f : \mathbb{R} \to \mathbb{R}\) of the form
\[
f(x) = x\,\phi(x),
\]
where \(\phi : \mathbb{R} \to \mathbb{R}\) is a smooth gating function.
For GELU, \(\phi(x)\) is the cumulative distribution function of the standard
normal distribution, while for Swish, \(\phi(x)\) is the sigmoid function.

Applied elementwise to an input \(\boldsymbol{X} \in \mathbb{R}^{t \times d}\),
this nonlinearity can be written as
\[
F(\boldsymbol{X}) = \boldsymbol{X} \odot \boldsymbol{\Phi}(\boldsymbol{X}),
\]
where \(\boldsymbol{\Phi}(\boldsymbol{X})\) denotes the elementwise application of \(\phi\).
Accordingly, the operator-valued map is given by
\[
L(\boldsymbol{X})(\boldsymbol{X}) = \boldsymbol{\Phi}(\boldsymbol{X}) \odot \boldsymbol{X},
\]
which corresponds directly to form~(II) in
Equation~\ref{eq:layer_representation}.
Substituting this expression into the layer definition
(Equation~\ref{eq:vit_layer}) recovers the standard GELU and Swish activations.

\subsubsection{Residual connections}
Residual connections combine the input with the output of a transformation
branch.
Let \(H : V_{\mathrm{in}} \to V_{\mathrm{in}}\) denote a residual branch that
admits a representation consistent with our framework, i.e.,
\[
H(\boldsymbol{X}) = L_H(\boldsymbol{X})(\boldsymbol{X}) + \boldsymbol{B}_H,
\]
or a composition of such transformations.
The corresponding residual layer is given by
\[
F(\boldsymbol{X}) = \boldsymbol{X} + H(\boldsymbol{X}).
\]
Substituting the representation of \(H\) yields
\[
F(\boldsymbol{X})
=
\left(\boldsymbol{I} + L_H(\boldsymbol{X})\right)(\boldsymbol{X})
+ \boldsymbol{B}_H,
\]
which shows that residual connections extend the effective transformation by
adding the identity operator.
Accordingly, the residual layer admits an effective operator
\[
L_F(\boldsymbol{X}) = \boldsymbol{I} + L_H(\boldsymbol{X})
\]

\subsection{Effective transformation matrix}
\label{app:effective_weight}

In this section, we show that for both operator forms~(I) and~(II) introduced in
Equation~\ref{eq:layer_representation}, the effective transformation
\(L(\boldsymbol{X})\) admits an input-dependent matrix
representation acting on the vectorized layer input.

\paragraph{Matrix representation.} Let \(V_{\mathrm{in}}\) and \(V_{\mathrm{out}}\) denote the input
and output vector spaces of a layer, and fix bases
\(\mathcal{B}_{\mathrm{in}}\) and \(\mathcal{B}_{\mathrm{out}}\) for these spaces.
For any operator-valued map
\(L(\boldsymbol{X}) \in \mathcal{L}(V_{\mathrm{in}}, V_{\mathrm{out}})\),
the linear operator \(L(\boldsymbol{X})\) admits a (generally input-dependent)
matrix representation
\(\boldsymbol{W}_L(\boldsymbol{X})\) with respect to these bases, such that
\[
[L(\boldsymbol{X})(\boldsymbol{X})]_{\mathcal{B}_{\mathrm{out}}}
=
\boldsymbol{W}_L(\boldsymbol{X})\,
[\boldsymbol{X}]_{\mathcal{B}_{\mathrm{in}}}.
\]

\paragraph{ViT vector spaces.} In the case of Vision Transformer layers, we identify
\(V_{\mathrm{in}} \cong \mathbb{R}^{t \times d_{\mathrm{in}}}\) and
\(V_{\mathrm{out}} \cong \mathbb{R}^{t \times d_{\mathrm{out}}}\), and represent
these spaces via vectorization. We take \(\mathcal{B}_{\mathrm{in}}\) and \(\mathcal{B}_{\mathrm{out}}\) to be the
standard bases of \(\mathbb{R}^{t d_{\mathrm{in}}}\) and
\(\mathbb{R}^{t d_{\mathrm{out}}}\), respectively, so that:
\[
[L(\boldsymbol{X})(\boldsymbol{X})]_{\mathcal{B}_{\mathrm{out}}}
=
\mathrm{vec}\!\left(L(\boldsymbol{X})(\boldsymbol{X})\right).
\]

\paragraph{Form~(I).}
For operator form~(I) (see Equation~\ref{eq:layer_representation}), by applying the vectorization operator, we obtain:
\[
\mathrm{vec}\!\left(L(\boldsymbol{X})(\boldsymbol{X})\right)
=
\left(
\boldsymbol{W}_d^\top \otimes \boldsymbol{W}_t(\boldsymbol{X})
\right)\mathrm{vec}(\boldsymbol{X}).
\]
where $\otimes$ denotes a Kronecker product. Thus, under the bases induced by vectorization, the effective transformation
admits the matrix representation:
\[
\boldsymbol{W}_L(\boldsymbol{X})
=
\boldsymbol{W}_d^\top \otimes \boldsymbol{W}_t(\boldsymbol{X})
\in \mathbb{R}^{(t d_{\mathrm{out}}) \times (t d_{\mathrm{in}})}
\]
In particular, for self-attention layers,
\(\boldsymbol{W}_t(\boldsymbol{X})\) is given by the attention matrix and
\(\boldsymbol{W}_d\) encodes the value and output projections, yielding a
Kronecker-structured effective weight matrix.

\paragraph{Form~(II)}
For operator form~(II) (see Equation~\ref{eq:layer_representation}), vectorizing the elementwise product yields
\[
\mathrm{vec}\!\left(\boldsymbol{\Phi}(\boldsymbol{X}) \odot \boldsymbol{X}\right)
=
\mathrm{diag}\!\left(\mathrm{vec}(\boldsymbol{\Phi}(\boldsymbol{X}))\right)
\mathrm{vec}(\boldsymbol{X}),
\]
where \(\mathrm{diag}(\cdot)\) denotes the diagonal matrix formed from its vector
argument.
Hence the effective transformation admits the matrix representation
\[
\boldsymbol{W}_L(\boldsymbol{X})
=
\mathrm{diag}\!\left(\mathrm{vec}(\boldsymbol{\Phi}(\boldsymbol{X}))\right)
\in \mathbb{R}^{(t d_{\mathrm{in}}) \times (t d_{\mathrm{in}})},
\]

\subsection{Equivariant transformation}\label{app:equivariant}
In this section, we show that the proposed Reynolds-inspired operator
preserves equivariant attribution components and suppresses(is non-expansive) on
locally non-equivariant components.

\subsubsection{Reynolds Operator}\label{sec:reynolds}
Given a group action on a vector space, the Reynolds operator provides a
principled way to decompose each vector into components that are invariant and
non-invariant under the action.

\paragraph{Definition.} Let $G$ be a compact group acting linearly on a real vector space $V$
via a representation $\rho : G \to \mathcal{L}(V,V)$.
We assume that $G$ is equipped with a normalized Haar (uniform) measure $\mu$.
The Reynolds operator associated with this group action is defined as
the linear map $\mathcal{R} : V \to V$ given by:
\begin{equation}
\label{eq:reynolds_def}
    \mathcal{R}(\boldsymbol{x})
    := \int_G
\rho(g)(\boldsymbol{x})
\, d\mu(g)
\end{equation}

\paragraph{Projection onto invariant subspace.}
Defining the subspace of $G$-invariant elements as:
\begin{equation}\label{eq:projection}
    V^G := \{ \boldsymbol{x} \in V \mid \rho(g)(\boldsymbol{x}) = \boldsymbol{x}
    \;\; \forall g \in G \}
\end{equation}
the Reynolds operator $\mathcal{R} : V \rightarrow V$ satisfies
$\mathrm{Im}(\mathcal{R}) = V^G$ and $\mathcal{R} \circ \mathcal{R} = \mathcal{R}$,
and therefore is a projection onto $V^G$.

\paragraph{Vector space decomposition.}
As a consequence of projection from the previous paragraph, the vector space $V$ decomposes as a direct sum:
\begin{equation}\label{eq:g_decomposition}
    V = V^G \oplus \ker(\mathcal{R})
\end{equation}

Which decomposes each $\boldsymbol{x}\in V$ to $\boldsymbol{x} = \mathcal{R}(\boldsymbol{x}) + (\boldsymbol{x} - \mathcal{R}(\boldsymbol{x}))$. The component $\mathcal{R}(\boldsymbol{x}) \in V^G$ captures the $G$-invariant part of $\boldsymbol{x}$, and $(\boldsymbol{x} - \mathcal{R}(\boldsymbol{x})) \in \ker(\mathcal{R})$ contains all components that are not invariant under the group action.

\subsubsection{Suppression of non-equivariant components.}
In this section we show that the Reynolds-inspired operator defined in Eq.~\eqref{eq:equivariance}
suppresses (i.e., is non-expansive on) non-equivariant attribution components.

Let $V$ denote the Hilbert space of attribution maps equipped with the
inner product induced by the spatial domain, and let
$V^G \subset V$ be the subspace of $G$-equivariant elements under the action
\[
(\tau \cdot \boldsymbol{W}_L)(\boldsymbol{X})
:= [\tau^{-1}\circ\boldsymbol{W}_L\circ\tau](\boldsymbol{X}).
\]
According to Equation~\ref{eq:g_decomposition}, the ideal Reynolds operator associated with this action induces an orthogonal
decomposition
\[
\boldsymbol{W}_L
=
\boldsymbol{W}_L^{G}
+
\boldsymbol{W}_L^{\perp},
\]
where $\boldsymbol{W}_L^{G} \in V^G$ and
$\boldsymbol{W}_L^{\perp} \in (V^G)^{\perp}$.

Assuming that each spatial transformation $\tau$ acts unitarily on $V$
(i.e., preserves the inner product on the spatial domain),
the averaging operator from Eq.~\eqref{eq:averaging} is non-expansive:
\[
\|\boldsymbol{W}_L^{\mathrm{eq}}\|
\;\le\;
\|\boldsymbol{W}_L\|.
\]
Moreover, equivariant components are preserved exactly:
if $\boldsymbol{W}_L \in V^G$, then
$\boldsymbol{W}_L^{\mathrm{eq}} = \boldsymbol{W}_L$.

Thus, while the proposed operator does not define an exact Reynolds
projection (unless $\nu$ is the Haar measure), it preserves
the equivariant component of the attribution map and is
non-expansive on the non-equivariant component.

\subsection{Stochastic averaging as convolution}
\label{app:stochastic_convolution}

In this section, we show that the local averaging operation used in
Section~\ref{sec:low_pass} is equivalent to a convolution with the noise
distribution. Let $\boldsymbol{X} \in \mathbb{R}^d$ denote the (vectorized)
layer input and let
$\boldsymbol{W}^{\mathrm{eq}}_L : \mathbb{R}^d \to \mathbb{R}^m$ denote the
equivariant effective transformation (the argument below applies componentwise
when $m > 1$).

Let $\boldsymbol{\epsilon}$ be a random variable in $\mathbb{R}^d$ with
probability density $\mathcal{K} : \mathbb{R}^d \to \mathbb{R}_{\ge 0}$, $
\int_{\mathbb{R}^d} \mathcal{K}(\boldsymbol{z}) \,\mathrm{d}\boldsymbol{z} = 1$.
The low-pass filtered equivariant effective transformation is defined by
stochastic averaging,
\[
\tilde{\boldsymbol{W}}^{\mathrm{eq}}_L(\boldsymbol{X})
=
\mathbb{E}_{\boldsymbol{\epsilon}}
\left[
\boldsymbol{W}^{\mathrm{eq}}_L(\boldsymbol{X} + \boldsymbol{\epsilon})
\right]
=
\int_{\mathbb{R}^d}
\boldsymbol{W}^{\mathrm{eq}}_L(\boldsymbol{X} + \boldsymbol{\epsilon})
\,\mathcal{K}(\boldsymbol{\epsilon}) \,\mathrm{d}\boldsymbol{\epsilon}.
\]

We now rewrite this as a convolution via a change of variables. Setting $\boldsymbol{u} = \boldsymbol{X} + \boldsymbol{\epsilon}$,
$\boldsymbol{\epsilon} = \boldsymbol{u} - \boldsymbol{X}$, $\mathrm{d}\boldsymbol{\epsilon} = \mathrm{d}\boldsymbol{u}$,
to obtain:
\[
\tilde{\boldsymbol{W}}^{\mathrm{eq}}_L(\boldsymbol{X})
=
\int_{\mathbb{R}^d}
\boldsymbol{W}^{\mathrm{eq}}_L(\boldsymbol{u})\,
\mathcal{K}(\boldsymbol{u} - \boldsymbol{X}) \,\mathrm{d}\boldsymbol{u}.
\]
Which is the convolution of $\boldsymbol{W}^{\mathrm{eq}}_L$ with the kernel
$\mathcal{K}$:
\[
\tilde{\boldsymbol{W}}^{\mathrm{eq}}_L(\boldsymbol{X})
=
(\boldsymbol{W}^{\mathrm{eq}}_L \ast \mathcal{K})(\boldsymbol{X}).
\]

In the setting of Section~\ref{sec:low_pass}, $\mathcal{K}$ is chosen to be the
density of a zero-mean Gaussian distribution $\mathcal{N}(0,\Sigma)$, which
corresponds to convolution with a Gaussian kernel and yields the low-pass
filtering effect described in the main text.

\section{Additional Implementation Details}
\label{app:impl_details}

\paragraph{Models and checkpoints.}
All backbones are evaluated using the authors' publicly released \emph{pretrained} checkpoints from their official repositories. Unless stated otherwise, we do not fine-tune model weights for attribution evaluation; DAVE and all baselines are computed on frozen models under the same preprocessing and input resolution ($224{\times}224$).

\paragraph{DINO linear probing.}
For the self-supervised DINO ViT-B/16 backbone, we follow the standard linear-probing protocol on ImageNet-1k: we freeze the pretrained DINO backbone and train a linear classifier on top of the [CLS] representation using ImageNet-1k train labels, and evaluate on the ImageNet-1k validation set. We use a single linear layer trained with cross-entropy, SGD with momentum (0.9), and cosine learning-rate schedule (0.001) for 100 epochs with 128 batch-size. We use standard data augmentation for linear probing (random resized crop and horizontal flip) and otherwise default training settings from the authors' reference implementation~\cite{caron2021emerging}.

All attribution methods are then computed with respect to the resulting linear-probe classifier while keeping the DINO backbone frozen.

\begin{figure}[!h]
    \centering
    \includegraphics[width=1.0\linewidth]{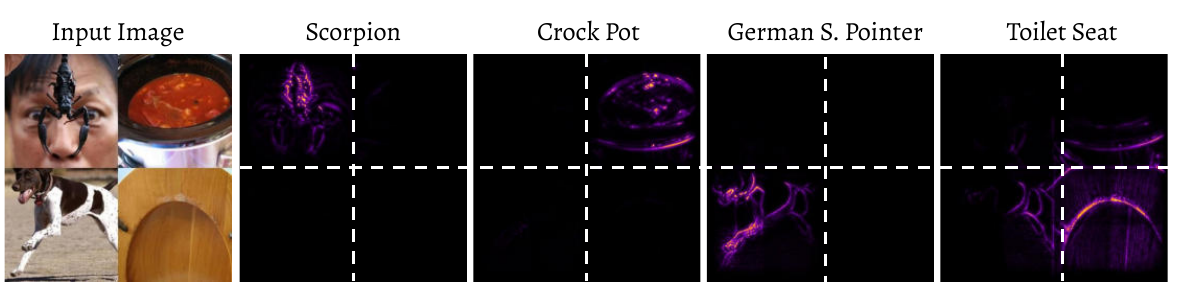}
    \caption{\textbf{GridPG Illustration.} 2x2 cells. }
    \vspace{-4pt}
    \label{fig:gridpg_illustration}
    \vspace{-4pt}
\end{figure}

\paragraph{Localization metrics.}
 For GridPG, we follow the popular protocol~\cite{boehle2021convolutional,zhang2018top}: we construct $500$ $2{\times}2$ grids from ImageNet-1k validation images of \emph{distinct} classes that are correctly classified with high confidence.\footnote{We use the same grid construction procedure across all attribution methods and models.} For each target class $i$, we compute the fraction of \emph{positive} attribution mass assigned to its corresponding grid cell. Let $A(p)$ be the attribution value at pixel $p$ and $A^+(p)=\max(A(p),0)$ its positive part. The per-cell score is
\begin{equation}
L_i=\frac{\sum_{p \in \text{cell}_i} A^+(p)}{\sum_{j=1}^{4} \sum_{p \in \text{cell}_j} A^+(p)},
\end{equation}
and the GridPG score is the average of $L_i$ over all constructed grids (see~\cref{fig:gridpg_illustration} for illustration).

Note: Samples with no-positive attributions in the entire grid are dropped from evaluation as is commonly done~\cite{boehle2021convolutional}.

\begin{figure}[!h]
    \centering
    \includegraphics[width=0.8\linewidth]{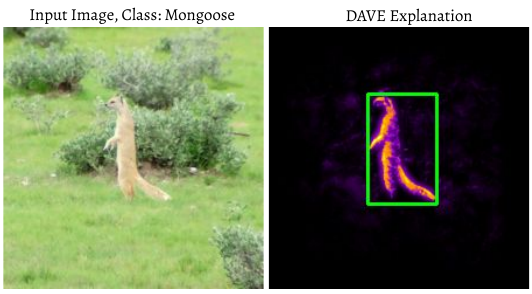}
    \caption{\textbf{EnergyPG Illustration.}}
    \vspace{-4pt}
    \label{fig:epg_illustration}
    \vspace{-4pt}
\end{figure}

For the EnergyPG, we follow the common practice~\cite{scorecam} and evaluate on the ImageNet-1k validation set using the official ILSVRC~\cite{imagenet} bounding-box annotations. Let $\Omega$ denote the ground-truth bounding box region for the image label (union if multiple boxes are provided) and $\bar{\Omega}$ its complement (see~\cref{fig:gridpg_illustration} for illustration). We report
\begin{equation}
\mathrm{EnergyPG}=\frac{\sum_{p \in \Omega} A^+(p)}{\sum_{p \in \Omega \cup \bar{\Omega}} A^+(p)}.
\end{equation}

\paragraph{Faithfulness via pixel deletion.}
We evaluate faithfulness with \textit{pixel deletion} curves following prior work~\cite{chefer2021transformer,boehle2021convolutional}. Given an attribution map, we rank pixels in \emph{increasing} order of attribution (least to most important) and progressively set the lowest-ranked pixels to zero. After each deletion step, we forward the modified image through the model and record the target-class probability. We plot the target probability versus the fraction of pixels removed.

\paragraph{Attribution post-processing.}
Unless stated otherwise, we evaluate attributions at the model input resolution. For methods that produce patch-level maps, we upsample to pixel space using bilinear interpolation. For all metrics, we use the positive attribution mass $A^+(p)$ as defined above.

\paragraph{Algorithm.}
We preserve the standard forward pass while enabling efficient evaluation of the effective transformation via a modified gradient computation.
Specifically, we reimplement each layer $F_i$ of the ViT according to Equation~\ref{eq:vit_layer}.
For each operator-valued map $L$ within a layer, we introduce a separate conditioning variable $\boldsymbol{c}_i$ that parameterizes the operator as $L(\boldsymbol{c}_i)$, while the input variable $\boldsymbol{x}_i$ is used to evaluate $L(\boldsymbol{c}_i)(\boldsymbol{x}_i)$.
Although $\boldsymbol{c}_i$ and $\boldsymbol{x}_i$ take identical values during the forward pass (see Algorithm~\ref{alg:conditioned-forward}), gradients through $\boldsymbol{c}_i$ are blocked during backpropagation.
This construction leaves the forward computation unchanged while allowing gradients with respect to the input $\boldsymbol{x}$ to ignore operator variation, enabling efficient computation of the effective transformation.

Using the conditioned forward pass, we compute DAVE attributions as described in Algorithm~\ref{alg:dave-attribution}.
At each step, we sample a spatial transformation and input noise to generate a perturbed input.
We then evaluate the class score using the conditioned forward pass and compute the effective transformation via a modified backward pass with respect to the input, which ignores the operator variation.
The resulting effective transformation matrix is mapped back to the original input space by applying the inverse spatial transformation and accumulated across samples.
Finally, the aggregated effective transformation is multiplied elementwise with the input, yielding the final DAVE attribution map.

\begin{algorithm}[tb]
   \caption{Conditioned forward pass}
   \label{alg:conditioned-forward}
\begin{algorithmic}
   \STATE {\bf Input:} input $\boldsymbol{x}$; layers $\{F_i\}_{i=1}^n$
   \STATE {\bf Output:} logits $\boldsymbol{z}$
   \STATE $\boldsymbol{x}_i \leftarrow \boldsymbol{x}$
   \FOR{$i = 1$ {\bf to} $n$}
      \STATE $\boldsymbol{c}_i \leftarrow \mathrm{detach}(\boldsymbol{x}_i)$ 
      \STATE $\boldsymbol{x}_i \leftarrow F_i(\text{input}=\boldsymbol{x}_i, \text{conditioning}=\boldsymbol{c}_i)$
   \ENDFOR
   \STATE $\boldsymbol{z} \leftarrow \boldsymbol{x}_i$
   \STATE {\bf return} $\boldsymbol{z}$
\end{algorithmic}
\end{algorithm}

\begin{algorithm}[tb]
   \caption{DAVE attribution}
   \label{alg:dave-attribution}
\begin{algorithmic}
   \STATE {\bf Input:} data sample $\boldsymbol{x}$, class label $y$, number of samples $T$
   \STATE {\bf Hyperparameters:} transformation distribution $\nu$, low-pass noise distribution $\mathcal{K}$
   \STATE {\bf Output:} attribution map $\boldsymbol{A}$
   \STATE $\tilde{\boldsymbol{W}}^{\mathrm{eq}}_L \leftarrow \boldsymbol{0}$
   \FOR{$t = 1$ {\bf to} $T$}
      \STATE $\tau \sim \nu,\quad \boldsymbol{\epsilon} \sim \mathcal{K}$
      \STATE $\boldsymbol{x}_t \leftarrow \tau(\boldsymbol{x} + \boldsymbol{\epsilon})$
      \STATE $s_t \leftarrow \textsc{ConditionedForward}(\boldsymbol{x}_t)[y]$
      \STATE $\boldsymbol{W}_t \leftarrow \nabla_{\boldsymbol{x}_t}s_t$  \COMMENT{ignores conditioning due to conditioned forward pass}
      \STATE $\tilde{\boldsymbol{W}}^{\mathrm{eq}}_L \leftarrow \tilde{\boldsymbol{W}}^{\mathrm{eq}}_L + \tau^{-1}(\boldsymbol{W}_t)$
   \ENDFOR
   \STATE $\boldsymbol{A} \leftarrow \left(\frac{1}{T} \tilde{\boldsymbol{W}}^{\mathrm{eq}}_L\right) \odot \boldsymbol{x}$
   \STATE {\bf return} $\boldsymbol{A}$
\end{algorithmic}
\end{algorithm}

\section{Qualitative results}
\label{app:qualitative}

\begin{figure*}
    \centering
    \includegraphics[width=1.0\textwidth]{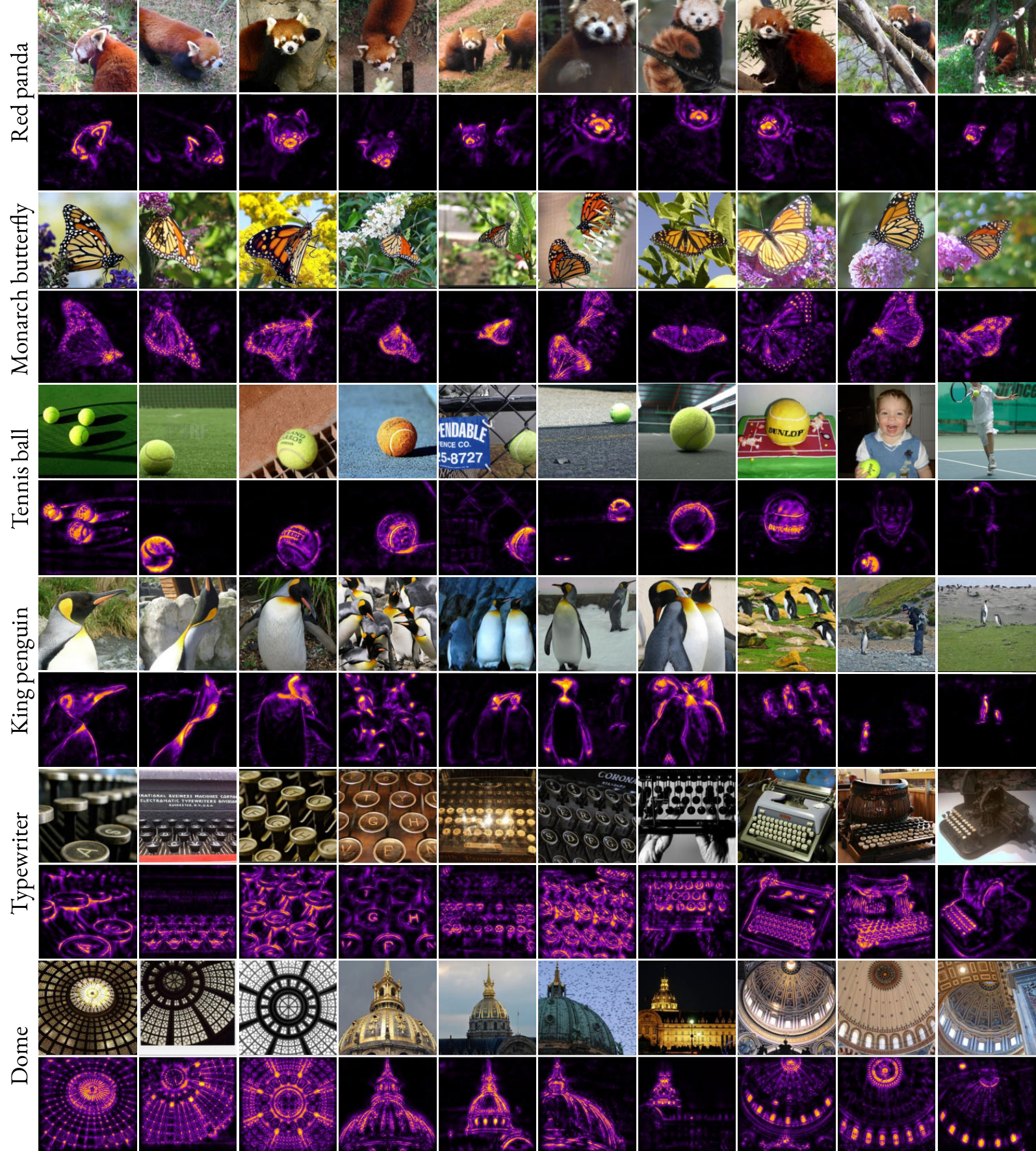}
    \caption{\textbf{Examples of class-consistent features detected by DAVE on DeiT-III-B/16-224.} For each ImageNet-1K class (rows), we show multiple validation images (top) and the corresponding DAVE attribution maps (bottom). Across diverse instances, DAVE consistently highlights recurring, semantically meaningful class cues (e.g., characteristic parts, textures, and contours), while adapting to changes in pose, scale, and background.}

    \label{fig:app_class_consist}
\end{figure*}

\begin{figure*}
    \centering
    \includegraphics[width=0.95\textwidth]{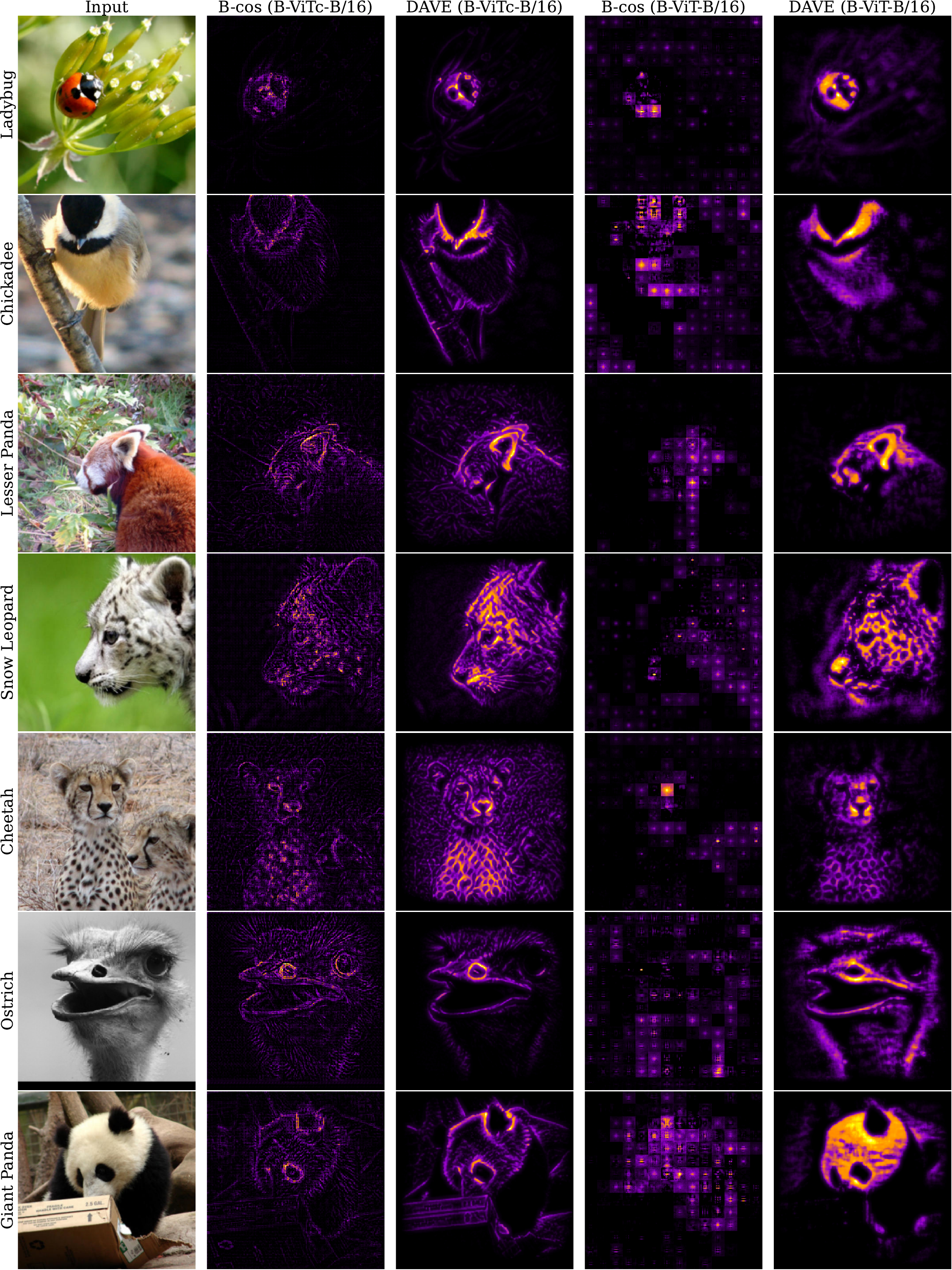}
    \caption{\textbf{DAVE explanations on inherently interpretable \bcos ViTs}: Our (DAVE) explanations visually highlight object centric features precisely especially for \bcos ViTs trained without a convolutional stem, showing robustness of the proposed method to architectural variations (see col. 4 vs col. 5) Even for \bcos ViTs with a convolutional stem, the DAVE attributions seem to highlight similar regions as the \bcos ones (col. 2 vs col 3), and also improving upon the \bcos localization performance.}
    \label{fig:dave_vs_bcos}
\end{figure*}

\begin{figure*}[t]
  \centering

  \begin{subfigure}[t]{0.7\textwidth}
    \centering
    \includegraphics[width=\linewidth]{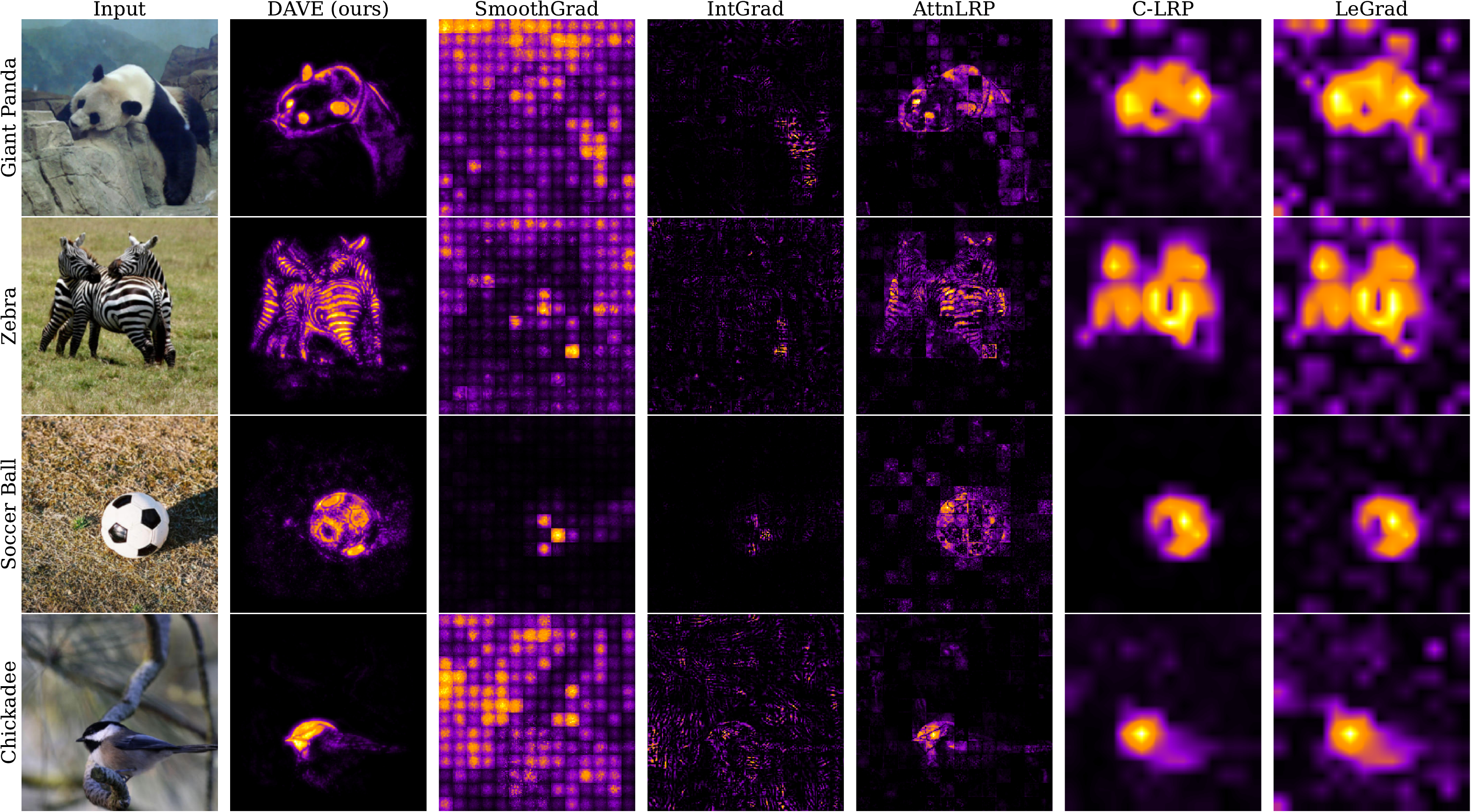}
    \caption{Deit-III-B/16-224 Model}
    \label{fig:attr_a_dave_vs_rest}
  \end{subfigure}


  \begin{subfigure}[t]{0.7\textwidth}
    \centering
    \includegraphics[width=\linewidth]{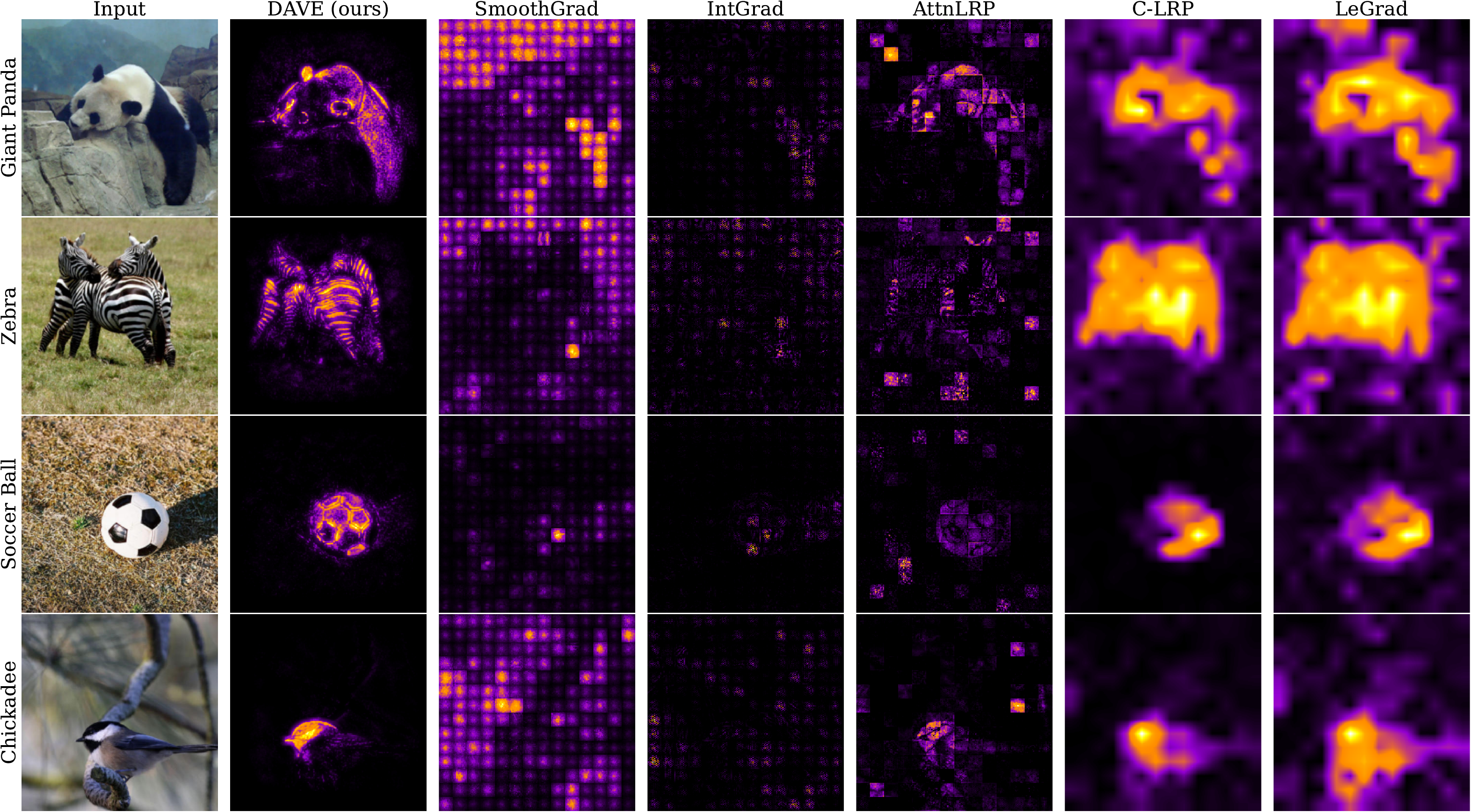}
    \caption{Deit-B/16-224 Model}
    \label{fig:attr_b_dave_vs_rest}
  \end{subfigure}


  \begin{subfigure}[t]{0.7\textwidth}
    \centering
    \includegraphics[width=\linewidth]{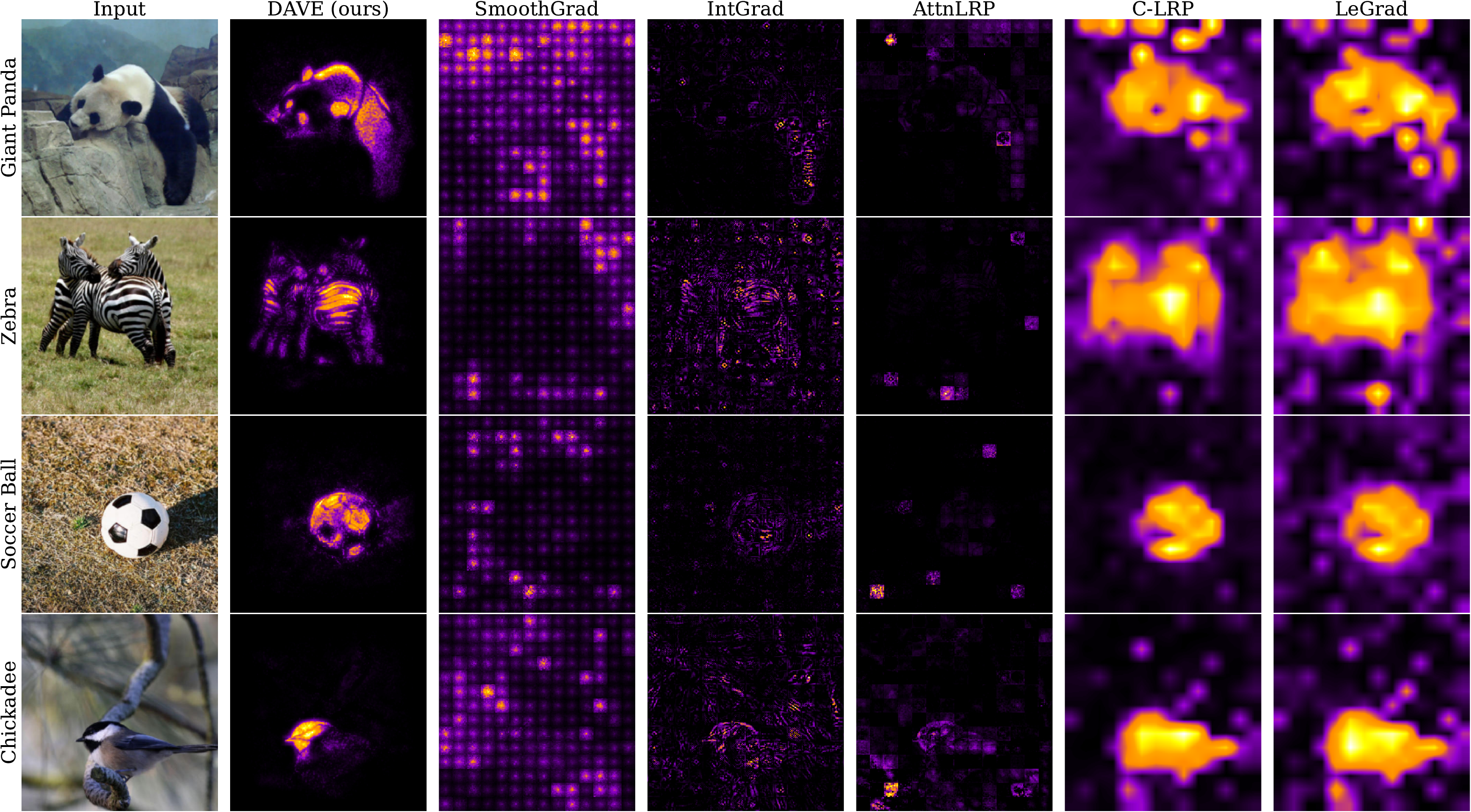}
    \caption{ViT-B/16-224 Model}
    \label{fig:attr_c_dave_vs_rest}
  \end{subfigure}

    \caption{\textbf{DAVE vs.\ post-hoc attribution methods across ViT backbones.} For ImageNet-1K validation examples (rows), we show the input image (left) and attribution maps produced by DAVE, SmoothGrad, Integrated Gradients, AttnLRP, C-LRP, and LeGrad (columns). Results are shown for (a) DeiT-III-B/16-224, (b) DeiT-B/16-224, and (c) ViT-B/16-224. Across backbones, DAVE yields sharper, more object-aligned and spatially coherent explanations with reduced patch-grid artifacts compared to prior methods.}

  \label{fig:attr_dave_vs_rest_models}
\end{figure*}

In this section, we provide qualitative analyses of DAVE attributions on ImageNet-1K validation samples. We first compare DAVE explanations across ViT-B/16-224, DeiT-B/16-224, and DeiT-III-B/16-224, highlighting when different models rely on shared versus distinct but semantically meaningful cues (\cref{fig:attr_three}). We then illustrate class-consistent attributions produced by DAVE, where recurring class-relevant features are highlighted across different images (\cref{fig:app_class_consist}). Next, we compare DAVE with the inherent explanations of B-cos ViTs, showing that DAVE produces sharper and more object-aligned maps while remaining robust to architectural variations such as the presence or absence of a convolutional stem (\cref{fig:dave_vs_bcos}). Finally, we provide qualitative comparisons of DAVE against prior post-hoc attribution methods, both on a self-supervised DINO ViT-B/16-224 model (\cref{fig:dave_vs_posthoc_dino}) and consistently across multiple ViT backbones (Deit-III-B/16, Deit-B/16, ViT-B/16)(\cref{fig:attr_dave_vs_rest_models}).

\subsection{Model comparison}
\begin{figure*}[t]
  \centering

  \begin{subfigure}[t]{0.8\textwidth}
    \centering
    \includegraphics[width=\linewidth]{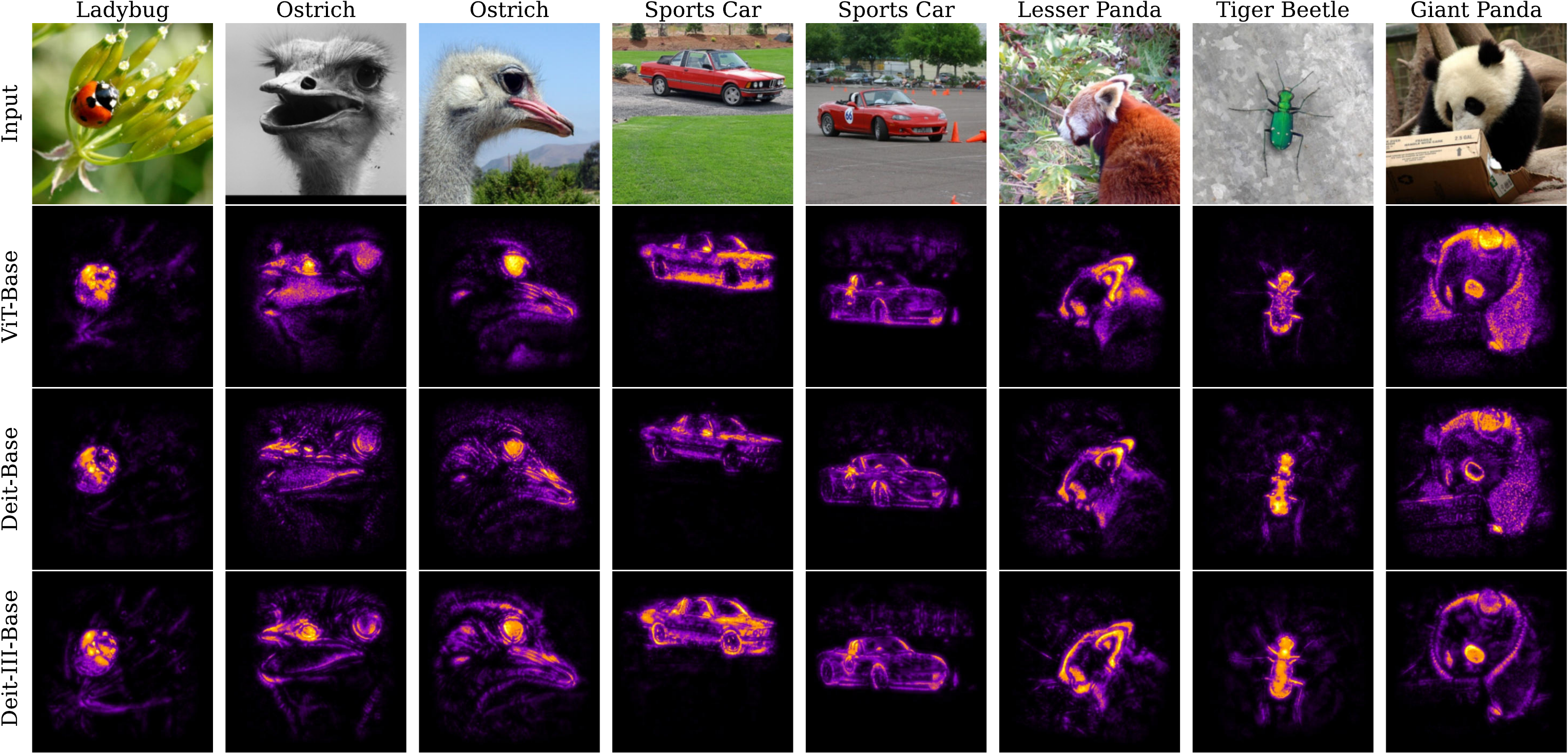}
    \caption{Set 1.}
    \label{fig:attr_a}
  \end{subfigure}


  \begin{subfigure}[t]{0.8\textwidth}
    \centering
    \includegraphics[width=\linewidth]{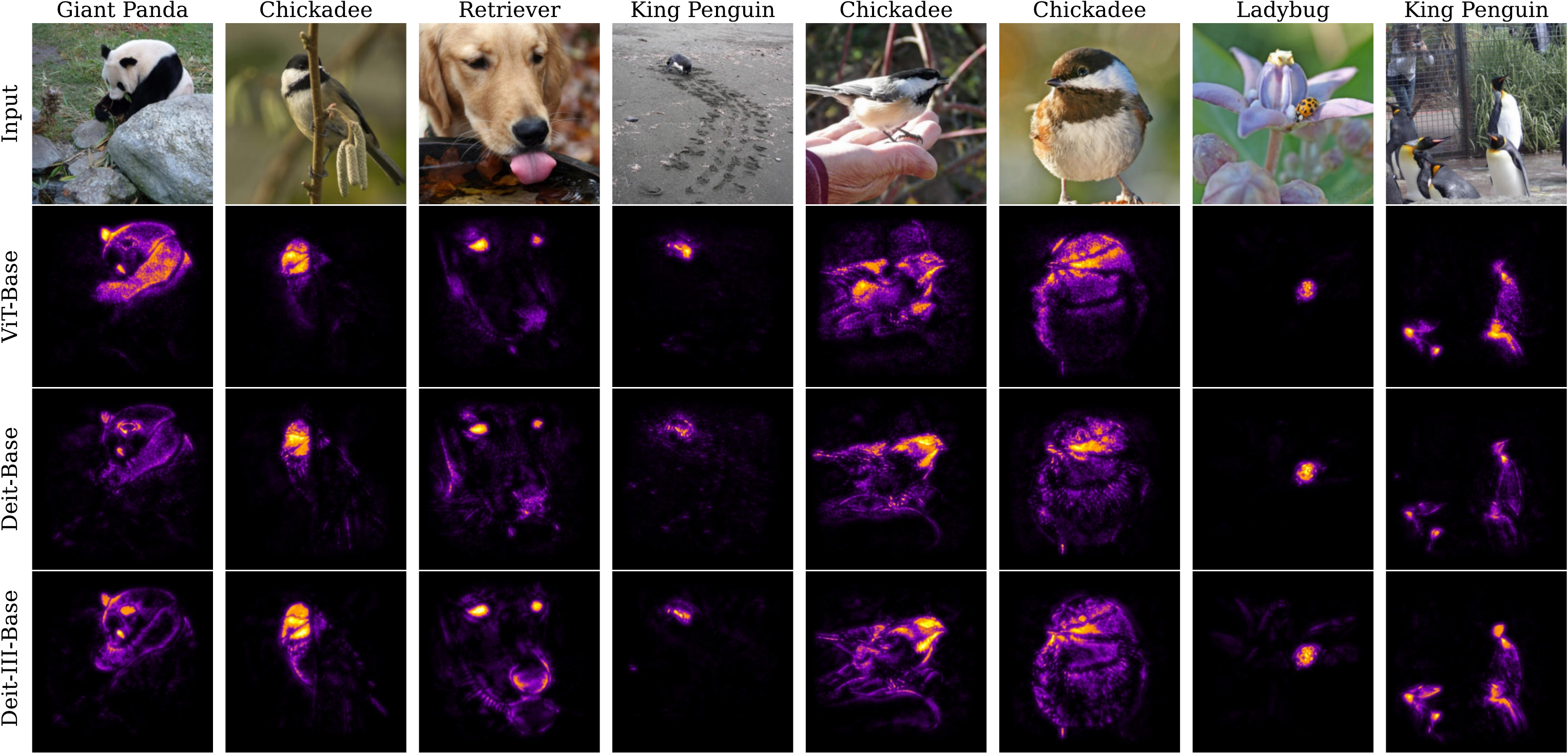}
    \caption{Set 2.}
    \label{fig:attr_b}
  \end{subfigure}


  \begin{subfigure}[t]{0.8\textwidth}
    \centering
    \includegraphics[width=\linewidth]{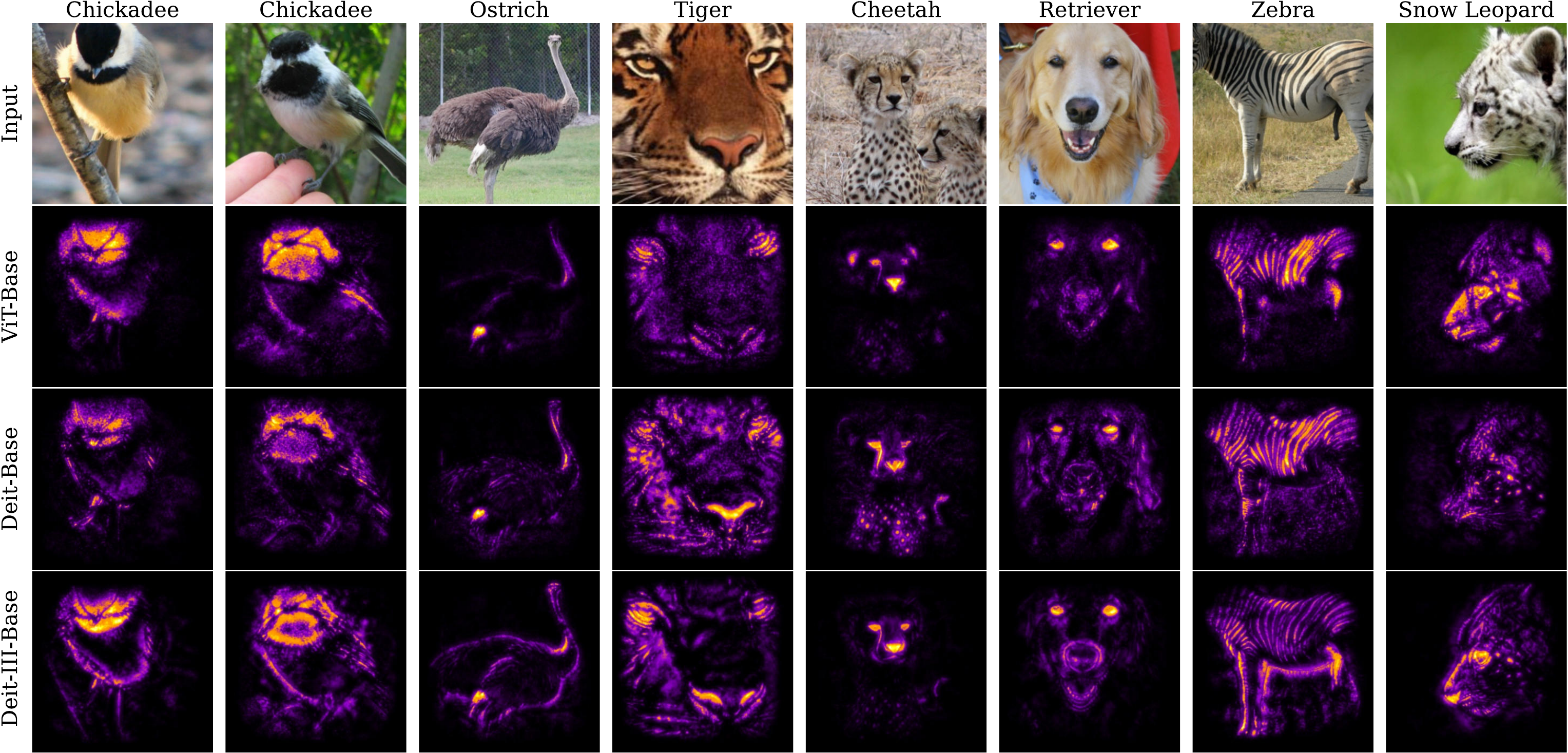}
    \caption{Set 3.}
    \label{fig:attr_c}
  \end{subfigure}

    \caption{\textbf{DAVE explanations across ViT backbones.} For ImageNet-1K validation examples (columns), we show the input image (top) and DAVE attribution maps produced by ViT-B/16-224, DeiT-B/16-224, and DeiT-III-B/16-224 (rows). Across samples, the three models often highlight similar object-centric, class-relevant regions, while in some cases emphasizing different but semantically meaningful cues, reflecting differences in learned feature representations.}

  \label{fig:attr_three}
\end{figure*}

We qualitatively compare attributions produced by DAVE on ViT-B/16-224, DeiT-B/16-224, and DeiT-III-B/16-224 for ImageNet-1K evaluation samples (see Figure~\ref{fig:attr_three}). While for some samples, the models attend to similar regions and features (indicating shared representations of class-relevant cues), for others, the models emphasize different regions. However, in both cases these regions remain semantically associated with the target object, suggesting alternative but valid feature representations across architectures.

\begin{figure*}
    \centering
    \includegraphics[width=0.93\textwidth]{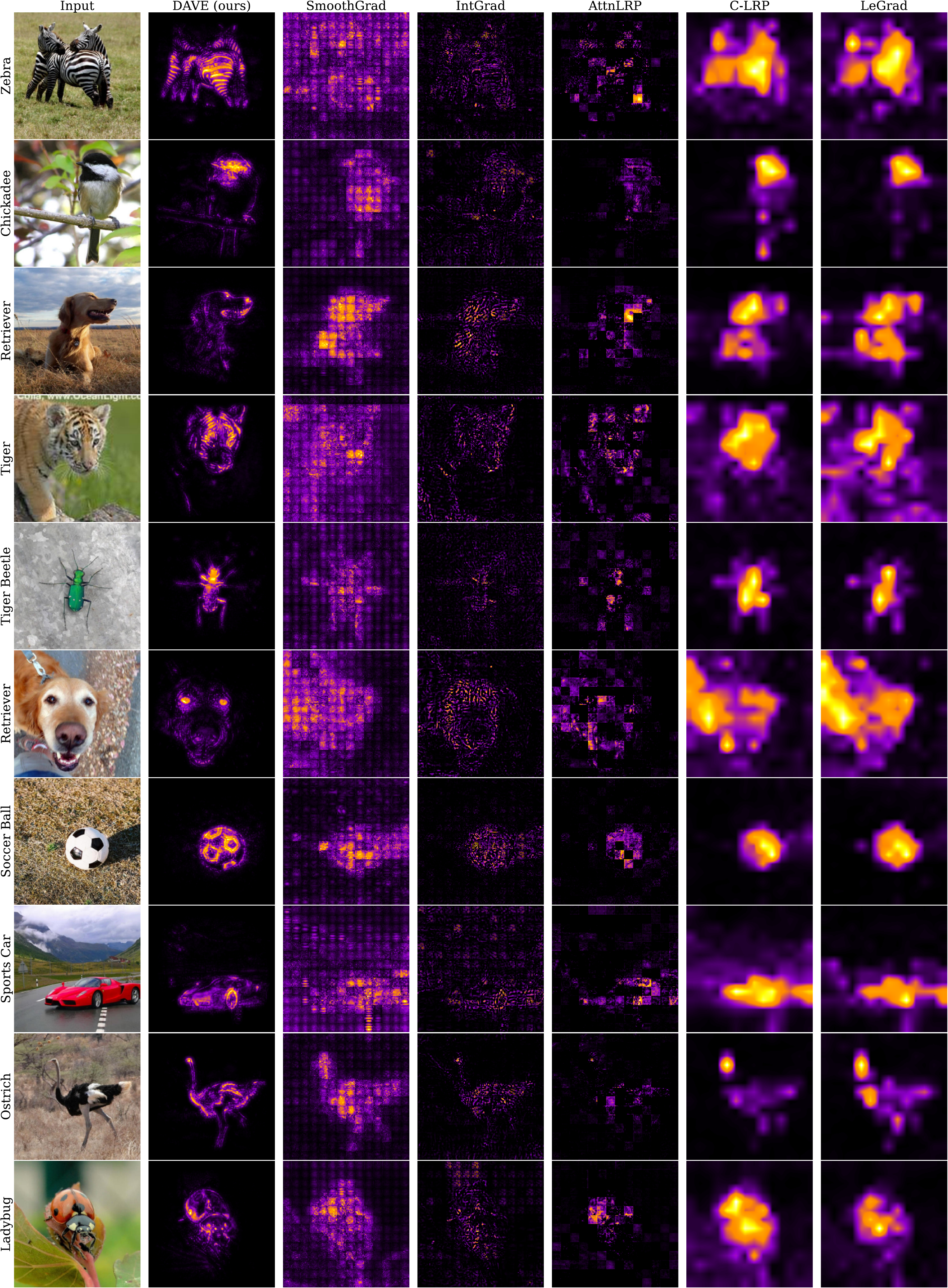}
  \caption{\textbf{DAVE vs.\ post-hoc attribution methods on DINO ViT-B/16-224.} For ImageNet-1K validation examples (rows), we show the input image (left) and attribution maps produced by DAVE, SmoothGrad, Integrated Gradients, AttnLRP, C-LRP (Chefer-LRP), and LeGrad (columns). DAVE yields sharper, more object-aligned and spatially coherent explanations with reduced patch-grid artifacts compared to prior methods.}
    \label{fig:dave_vs_rest_dino}
\end{figure*}

\subsection{Class consistency}
In the qualitative analysis, we observe class-consistent feature attributions produced by DAVE, where highlighted regions correspond to semantically meaningful and recurring features associated with the predicted class across different images (see Fig.~\ref{fig:app_class_consist}).

\subsection{Comparison with inherently interpretable models}
\label{app:results_bcos}

Figure~\ref{fig:dave_vs_bcos} compares DAVE to the inherent explanations of B-cos ViTs across ImageNet-1k validation examples. While B-cos explanations can be noisy---especially for B-cos ViTs trained without a convolutional stem (B-ViT-C)---DAVE yields sharper, more object-aligned attributions that better suppress background noisy responses. Notably, DAVE preserves class-relevant structure (e.g., eyes, contours, and distinctive textures) and produces more spatially coherent maps across both B-ViT-C and the convolutional-stem B-ViT variant, consistent with the localization improvements reported in Table~\ref{tab:pg_bcos_pct}.

\subsection{Comparison with post-hoc attributions on DINO}
\label{app:results_dino_posthoc}

Figure~\ref{fig:dave_vs_rest_dino} compares DAVE to common post-hoc attribution methods on a self-supervised DINO ViT-B/16-224 model~\cite{caron2021emerging}, including SmoothGrad, Integrated Gradients, AttnLRP, Chefer-LRP, and LeGrad.
Across examples, several baselines exhibit pronounced patch-grid artifacts, diffuse responses, or fragmented saliency that extends into the background.
In contrast, DAVE consistently yields sharper, more spatially coherent attributions that concentrate on object-centric and class-relevant structures (e.g., contours and distinctive parts), while reducing structured artifacts introduced by tokenization and attention routing.
These observations align with our quantitative localization results in the main paper and support DAVE’s ability to produce high-resolution, visually interpretable attributions on ViT based models.

\subsection{Comparison with post-hoc attributions across ViT backbones}
\label{app:results_dave_vs_rest_models}

Figure~\ref{fig:attr_dave_vs_rest_models} compares DAVE to common post-hoc attribution methods across three ViT backbones (DeiT-III-B/16-224, DeiT-B/16-224, and ViT-B/16-224). Across all backbones, several baselines exhibit structured patch-grid artifacts (notably SmoothGrad), fragmented responses, or diffuse saliency that extends beyond the target object. In contrast, DAVE consistently produces sharper and more spatially coherent maps that concentrate on object-centric and class-relevant structures (e.g., animal contours and distinctive parts), while suppressing background responses. The examples suggest that DAVE’s qualitative advantages are not tied to a specific ViT variant, but persist across different transformer backbones and training recipes.

\section{Quantitative results}
In this section we provide additional quantitative results complementing the main paper. We report extended results on the pixel deletion benchmark, and provide further analysis supporting the design choices used in DAVE attribution. Specifically, we include convergence analysis of the iterative attribution estimates, as well as sensitivity analyses with respect to input rotations and additive input noise.

\begin{figure*}[t]
  \centering

  \begin{minipage}[t]{0.49\textwidth}
    \centering
    \includegraphics[width=\linewidth]{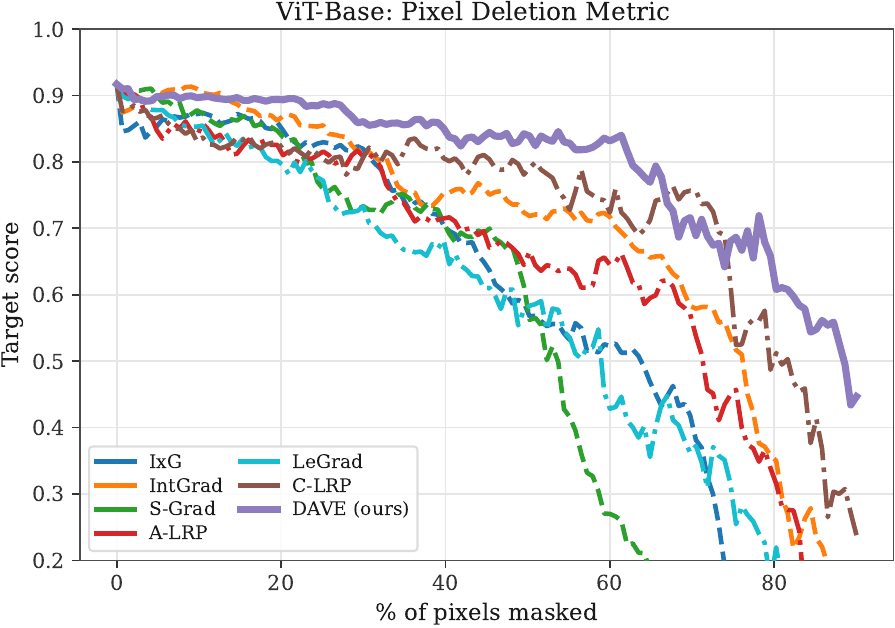}
  \end{minipage}\hfill
  \begin{minipage}[t]{0.49\textwidth}
    \centering
    \includegraphics[width=\linewidth]{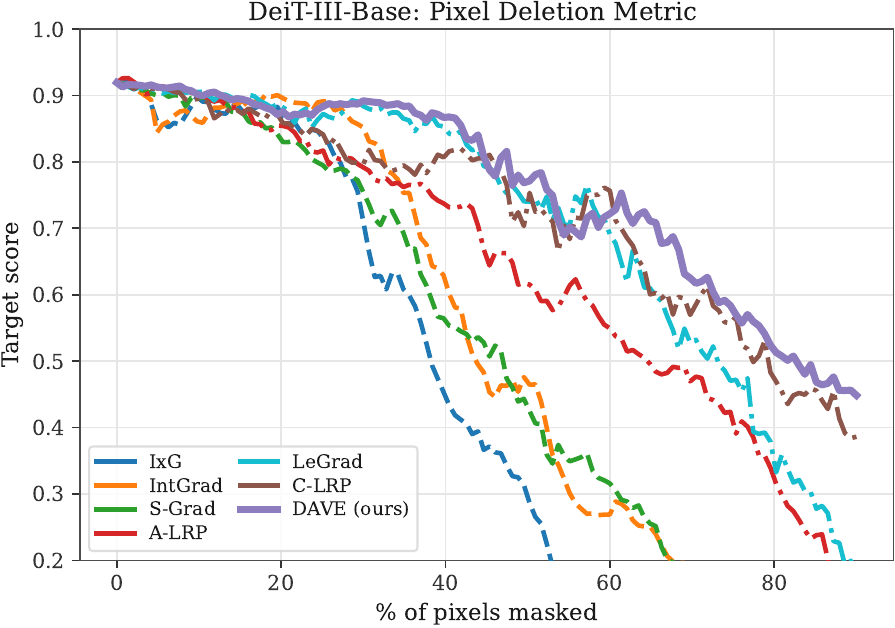}
  \end{minipage}

  \vspace{3mm}

  \begin{minipage}[t]{0.49\textwidth}
    \centering
    \includegraphics[width=\linewidth]{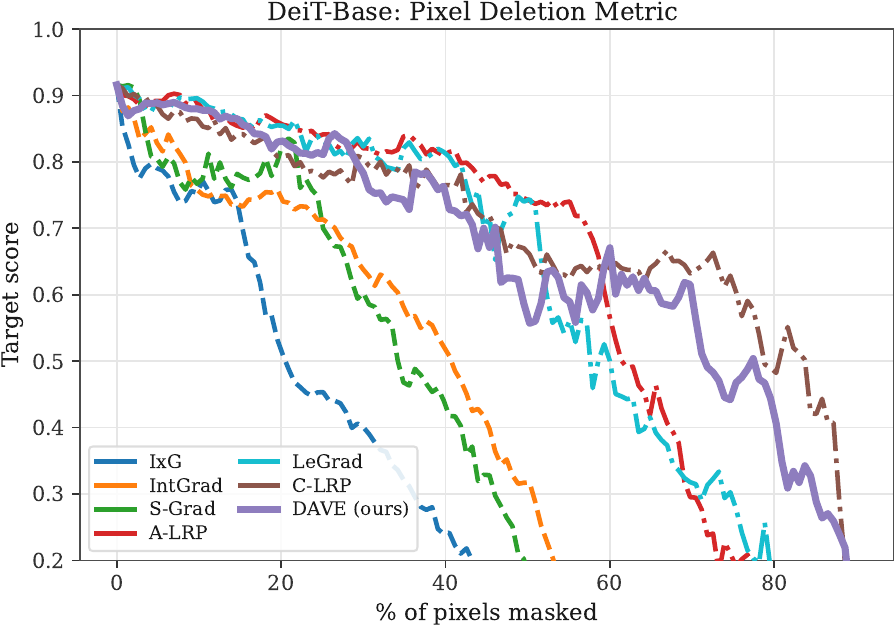}
  \end{minipage}\hfill
  \begin{minipage}[t]{0.49\textwidth}
    \centering
    \includegraphics[width=\linewidth]{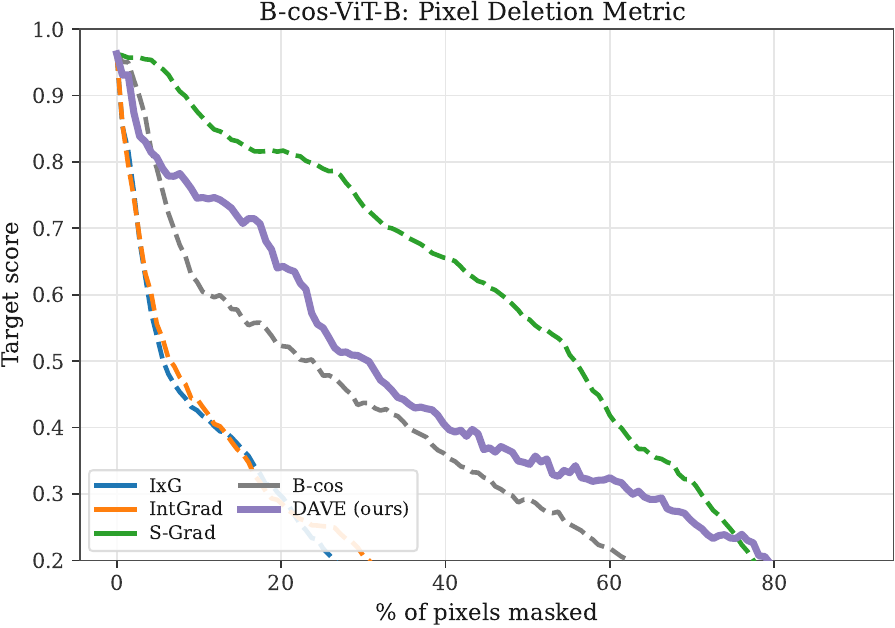}
  \end{minipage}

  \caption{\textbf{Pixel Deletion (\%)} for ViT-B/16 (row 1, col. 1), DeitIII-B/16 (row 1, col. 2), Deit-B/16 (row 2, col. 1) and B-cos-ViT-B/16 (row 2, col. 2) across attribution methods. Each curve corresponds to an attribution method; DAVE (ours) is highlighted in purple.}
  \label{fig:pixel_del_appendix}
\end{figure*}

\subsection{Ablation and Analysis}\label{app:ablation_analysis}

We provide additional analysis supporting the design choices used in DAVE attribution. We focus on three aspects: (i) convergence of the iterative attribution estimates, (ii) model sensitivity to input rotations, and (iii) model sensitivity to additive input noise. All analyses are conducted on ViT-B/16-224, DeiT-B/16-224, and DeiT-III-B/16-224, using 6{,}000 images randomly sampled from the ImageNet-1K validation set.

\begin{figure*}
    \centering
    \includegraphics[width=1.0\textwidth]{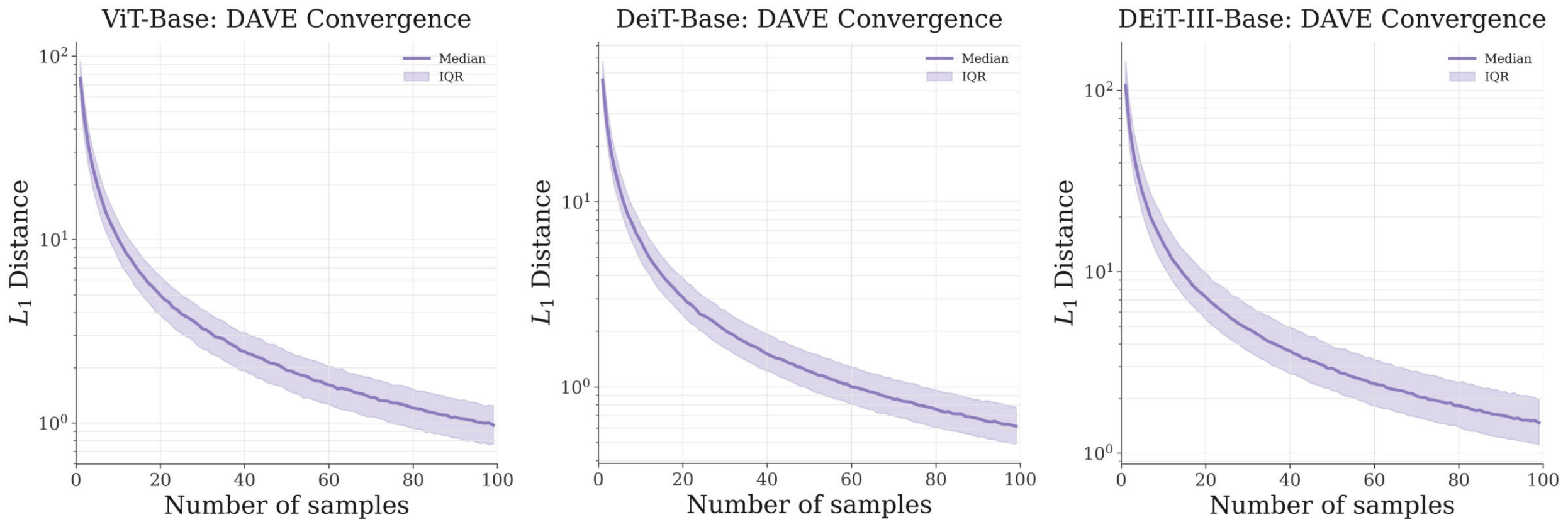}
    \caption{
\textbf{Convergence analysis of DAVE for ViT-B/16-224, DeiT-B/16-224, and DeiT-III-B/16-224.}
Convergence is measured as the summed $L_1$ distance between cumulative average attributions at successive averaging steps.
Results are reported as median and interquartile range across images and visualized on a logarithmic scale.
All models exhibit monotonically decreasing convergence, reaching values on the order of $10^{0}$ after approximately 100 steps.
}
    \label{fig:abl_converge}
\end{figure*}

\paragraph{Convergence Analysis.}
We analyze convergence of DAVE by measuring the summed $L_1$ distance between cumulative average attributions at successive averaging steps, which quantifies the marginal contribution of additional neighborhood samples (see Figure~\ref{fig:abl_converge}). For each step, results are summarized using the median and interquartile range across images and visualized on a logarithmic scale. Across all considered models, the convergence measure decreases rapidly, reaching values on the order of $10^{0}$ after approximately 100 steps. In practice, we observe no significant improvement in the evaluation metrics used in the paper beyond 50 steps, and therefore adopt this setting in the main experiments.

\paragraph{Rotation Sensitivity.}
We assess sensitivity to input rotations by measuring the absolute change in softmax probabilities of a target class relative to the unrotated input over a range of rotation angles. For each rotation angle, results are summarized using the median and interquartile range across the evaluated images.

\begin{figure*}
    \centering
    \includegraphics[width=1.0\textwidth]{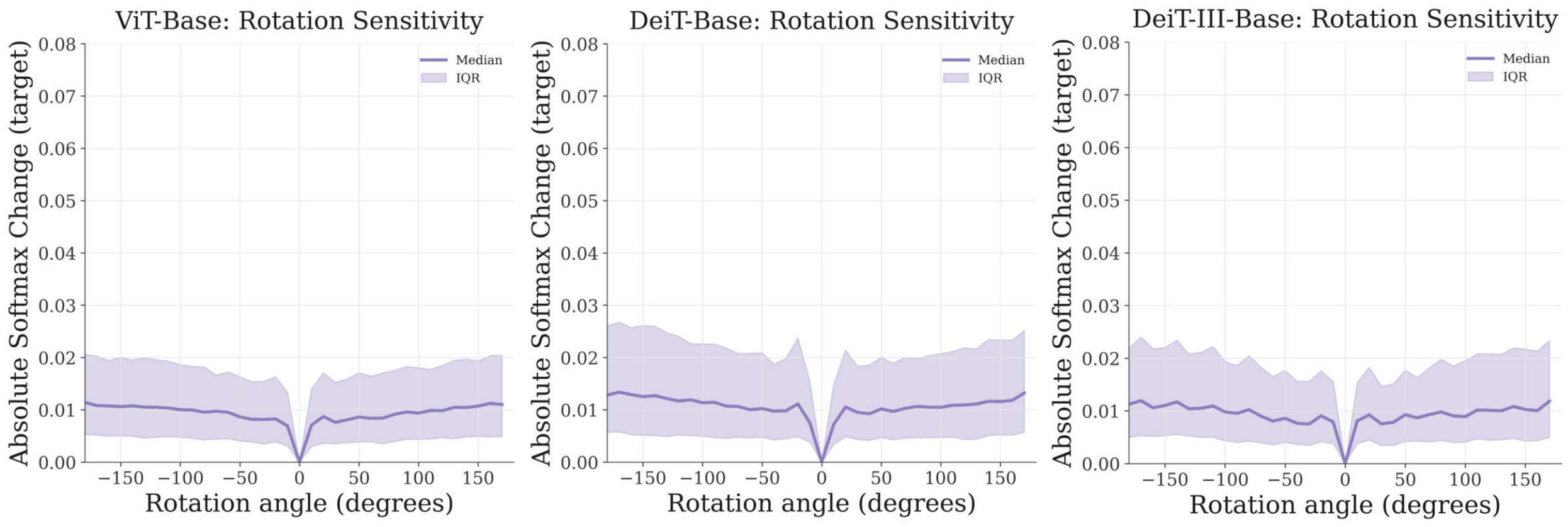}
    \caption{
\textbf{Rotation sensitivity analysis for ViT-B/16-224, DeiT-B/16-224, and DeiT-III-B/16-224.}
Sensitivity is measured as the absolute change in softmax probabilities relative to the unrotated input over rotation angles from $-180^\circ$ to $180^\circ$.
Results are summarized using the median and interquartile range across images.
All models exhibit relatively stable behavior under rotations, with smaller probability changes within the interval $(-20^\circ, 20^\circ)$.
}
    \label{fig:abl_rot}
\end{figure*}

All considered models exhibit relatively stable behavior under rotations over a broad range of angles (see Figure~\ref{fig:abl_rot}). Notably, rotations within $(-20^\circ, 20^\circ)$ result in smaller probability changes across models, and we therefore use this interval to define the attribution neighborhood.

\begin{figure*}
    \centering
    \includegraphics[width=1.0\textwidth]{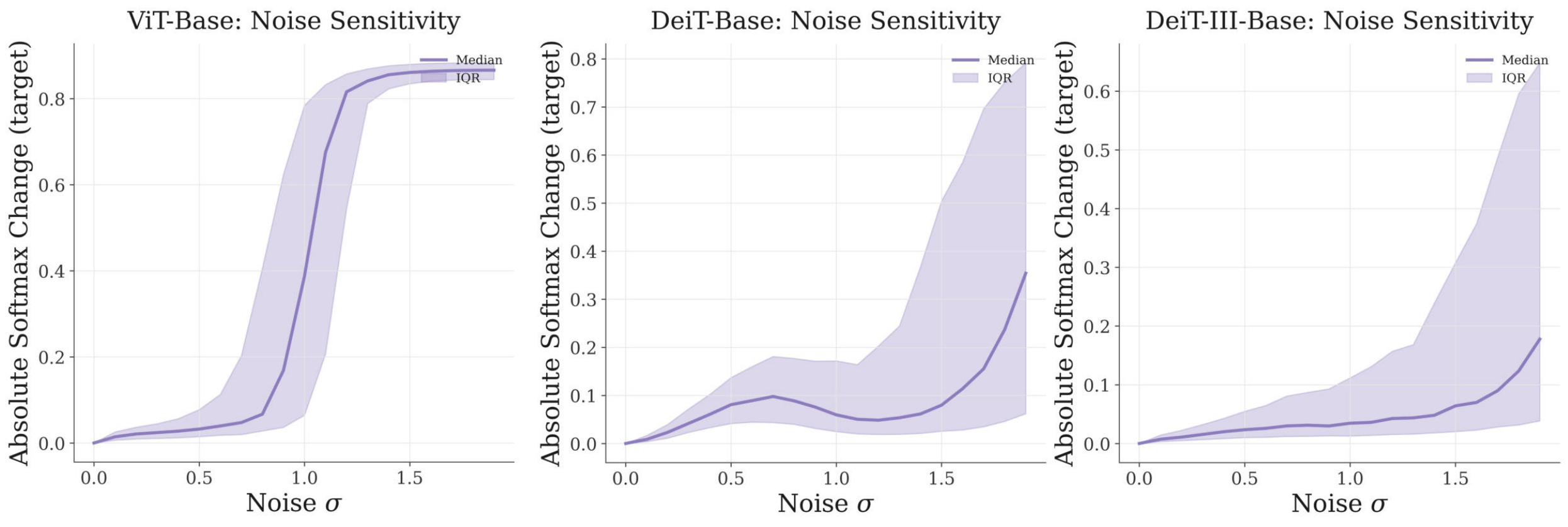}
    \caption{
\textbf{Noise sensitivity analysis for ViT-B/16-224, DeiT-B/16-224, and DeiT-III-B/16-224.}
Sensitivity is measured as the absolute change in softmax probabilities relative to the unperturbed input under additive Gaussian noise with standard deviation $\sigma \in [0, 2]$.
Results are summarized using the median and interquartile range across images.
}
    \label{fig:abl_noise}
\end{figure*}

\paragraph{Noise Sensitivity.}
We assess sensitivity to additive input noise using the same metric as for rotation analysis, measuring the absolute change in softmax probabilities relative to the unperturbed input. Gaussian noise with standard deviation $\sigma \in [0, 2]$ is added to the input, and results are summarized using the median and interquartile range across images for each noise level. We observe model-dependent sensitivity to noise: ViT-B/16-224 exhibits larger probability changes as noise increases, whereas DeiT-III-B/16-224 remains comparatively more stable (see Figure~\ref{fig:abl_noise}). 
This analysis is used to guide the selection of appropriate noise scales for effective weight smoothing (Section~\ref{sec:low_pass}) in subsequent experiments.


\subsection{DAVE Setup}\label{app:setup}

\paragraph{Transformations.}
For spatial transformations, we apply random horizontal flips with probability 0.5, small in-plane rotations uniformly sampled from $(-20^\circ, 20^\circ)$, and small wrapped translations with offsets up to 0.1 of the image extent along both spatial axes.

\paragraph{Noise injection.}
In Section~\ref{sec:low_pass}, we introduce low-pass filtering via expectation under Gaussian perturbations, corresponding to single-scale Gaussian smoothing.
While this is sufficient for B-cos models (using additive noise with $\sigma=0.2$),
we empirically find that conventional architectures benefit from a more robust
variant.

Specifically, we employ a variance-preserving noise interpolation scheme: for each sample we draw $\boldsymbol{\epsilon} \sim \mathcal{N}(0, I)$ and sample $t \in [0,0.5]$, constructing
\[
\boldsymbol{x}_t = (1 - t)\boldsymbol{x} + \sqrt{1 - (1 - t)^2}\,\boldsymbol{\epsilon}.
\]
Sampling $t$ effectively averages responses across noise scales, yielding a multi-scale generalization of the low-pass filter, which we find more robust for conventional models.

\subsection{Additional Results}

\paragraph{Pixel deletion results.}
Figure~\ref{fig:pixel_del_appendix} extends the pixel deletion evaluation to DeiT-B/16 and B-cos-ViT-B/16 (bottom row), in addition to ViT-B/16 and DeiT-III-B/16 (top row). Across all backbones, DAVE exhibits among the flattest deletion curves, indicating strong stability under progressive pixel removal. Notably, on B-cos-ViT-B/16, DAVE remains competitive with the inherent B-cos explanations and improves over standard gradient-based baselines.




\end{document}